\documentclass[12pt]{article}
\usepackage{amsmath}
\usepackage{graphicx,psfrag,epsf}
\usepackage{enumerate}
\usepackage{natbib}
\usepackage{mathtools}
\usepackage{color}
\usepackage{algorithm}
\usepackage{algpseudocode}

\makeatletter
\def\BState{\State\hskip-\ALG@thistlm}
\makeatother

\newcommand{\blind}{0}

\newcommand{\floor}[1]{\lfloor #1 \rfloor}

\begin{document}

\def\spacingset#1{\renewcommand{\baselinestretch}%
{#1}\small\normalsize} \spacingset{1}


\if0\blind
{
  \title{\bf Treeging}
  \author{Gregory L. Watson$^1$\thanks{
    gwatson@ucla.edu}\hspace{.2cm}, Michael Jerrett$^2$, Colleen E. Reid$^3$ and
    Donatello Telesca$^1$ \\
    $^1$ Department of Biostatistics, University of California, Los~Angeles \\ 
    $^2$ Department of Environmental Health Sciences, University of California, Los~Angeles
    \\ 
    $^3$ Geography Department, University of Colorado Boulder}
  \maketitle
} \fi

\if1\blind
{
  \bigskip
  \bigskip
  \bigskip
  \begin{center}
    {\LARGE\bf Title}
\end{center}
  \medskip
} \fi

\bigskip
\begin{abstract}
 Treeging combines the flexible mean structure of regression trees with the covariance-based prediction strategy of kriging into the base learner of an ensemble prediction algorithm. In so doing, it combines the strengths of the two primary types of spatial and space-time prediction models: (1) models with flexible mean structures (often machine learning algorithms) that assume independently distributed data, and (2) kriging or Gaussian Process (GP) prediction models with rich covariance structures but simple mean structures. We investigate the predictive accuracy of treeging across a thorough and widely varied battery of spatial and space-time simulation scenarios, comparing it to ordinary kriging, random forest and ensembles of ordinary kriging base learners. Treeging performs well across the board, whereas kriging suffers when dependence is weak or in the presence of spurious covariates, and random forest suffers when the covariates are less informative. Treeging also outperforms these competitors in predicting atmospheric pollutants (ozone and PM$_{2.5}$) in several case studies. We examine sensitivity to tuning parameters (number of base learners and training data sampling proportion), finding they follow the familiar intuition of their random forest counterparts. We include a discussion of scaleability, noting that any covariance approximation techniques that expedite kriging (GP) may be similarly applied to expedite treeging. 
\end{abstract}

\noindent%

\spacingset{1.45}
\section{Introduction}
\label{sec:intro}

Spatial or space-time prediction of a quantity based on observations of it at other locations or times (or both) is a common problem in fields as diverse as environmental health, real estate and mining. Formally, the problem is to predict the value of a random field $Y(d)$ at a new spatial or space-time location $d_0 \in \mathcal{D}$ given $n$ observations of the field at other locations, $\mathbf{y}(\mathbf{d})~=~[y_1(d_1), ..., y_n(d_n)]'$, where $d$ is the spatial ($\mathcal{D}~\subset~\mathcal{R}^2$) or space-time ($\mathcal{D}~\subset~\mathcal{R}^3$) coordinates that index the field. Ancillary information in the form of covariates  $\mathbf{X}({d})~=~[X_1(d), ..., X_p(d)]$ is generally available, in which case the goal is to predict $Y(d_0) \mid \mathbf{X}(d_0), \mathbf{z}(\mathbf{d})$, where $\mathbf{z}(\mathbf{d})~=~\{\mathbf{y}(\mathbf{d}), \mathbf{X}(\mathbf{d})\}$ is the observed data. The conditional expectation $E[Y(d_0) \mid \mathbf{X}(d_0), \mathbf{z}(\mathbf{d})]$ is the usual predictive target, because it is the minimal mean squared error (MSE) prediction of $Y(d_0) \mid \mathbf{X}(d_0), \mathbf{z}(\mathbf{d})$. 

If we suppose $Y(d) \mid \mathbf{X}(d)$ is composed of a mean function $f$ of the covariates at $d$ plus an error $\epsilon(d)$ that is independent of $\mathbf{X}(d)$ and has expectation 0, i.e., $E[\epsilon(d)]=0$, then 
\begin{equation} \label{y_given_x}
  Y(d) \mid \mathbf{X}(d) = f[\mathbf{X}(d)] + \epsilon(d).
\end{equation}
The second moment of $\epsilon(d)$ encodes the spatial or space-time dependence between values of the random field. In particular, let $\Sigma[d_1,d_2] \coloneqq \text{Cov}[\epsilon(d_1),\epsilon(d_2)]$, where $\Sigma$ is a positive semidefinite function defining the covariance between $\epsilon(d_1)$ and $\epsilon(d_2)$ (and therefore between $Y(d_1)$ and $Y(d_2)$) for any $d_1,d_2~\in~\mathcal{D}$.  In this setting, the conditional predictive target is decomposed as the sum of the mean function and the expected conditional error,
\begin{equation} \label{eq:cond_exp}
  E[Y(d_0) \mid \mathbf{X}(d_0), \mathbf{z}(\mathbf{d})] = f[\mathbf{X}(d_0)] + E[\epsilon(d_0) \mid \mathbf{z}(\mathbf{d})].
\end{equation}
It will be useful to characterize alternative prediction techniques based on the assumptions they make regarding these two terms. 

The dominant traditional approaches to spatial and space-time prediction, especially for interpolation, use simple mean structures to model $f$ and rely primarily on exploiting the spatial or space-time dependence of the random field for prediction. The mean function is usually assumed to be linear in the covariates, i.e., $f[\mathbf{X}(d_0)] = \mathbf{X}(d_0)'\boldsymbol \beta$, where $\boldsymbol \beta \in \mathcal{R}^p$. 

There is a rich literature on covariance estimation in this context, ranging from simple parametric functions of the distance between points~\citep{cressie2015spatial,gelfand2010handbook} to sophisticated anisotropic functions~\citep{paciorek2006spatial,sang2012full}. Given an estimate of the covariance function, $\hat{\Sigma}$, prediction proceeds by computing the best linear unbiased predictor (BLUP), 
\begin{equation} \label{eq:blup}
  \hat{Y}(d_0) | \mathbf{X}(d_0), \mathbf{z}(\mathbf{d}) = \mathbf{X}(d_0)' \hat{\boldsymbol \beta} + \hat{\Sigma}_0 \hat{\Sigma}(\mathbf{d})^{-1} [\mathbf{y}(\mathbf{d}) - \mathbf{X}(\mathbf{d}) \hat{\boldsymbol \beta}], 
\end{equation}
where $\hat{\Sigma} \coloneqq \hat{\Sigma}(\mathbf{d}, \mathbf{d})$ is the $n \times n$ estimated covariance matrix of the observed data,  $\hat{\Sigma}_0 \coloneqq \hat{\Sigma}(d_0, \mathbf{d})$ is the estimated covariance of the prediction location with the locations of the observed data, and $\hat{\boldsymbol \beta} = [\mathbf{X}(\mathbf{d})' \hat{\Sigma}(\mathbf{d})^{-1} \mathbf{X}(\mathbf{d})]^{-1}\mathbf{X}(\mathbf{d})'\hat{\Sigma}(\mathbf{d})^{-1}\mathbf{y}(\mathbf{d})$. This prediction procedure is widely known as kriging after Danie Krige~\citep{krige1951statistical}, the mining engineer who proposed it, and is a special case of the BLUP for generalized least squares~\citep{goldberger1962best}. It is also the maximum a posteriori (MAP) prediction of $Y(d_0)$ of a Bayesian Gaussian process (GP) model with an uninformative prior on $\boldsymbol \beta$ and an equivalent covariance structure. 
The common types of kriging in the literature (ordinary, universal, with external drift, etc.) amount to differences in which if any covariates are included in $\mathbf{X}(d)$. We remain general in notation but primarily ensivision a universal kriging model with the coordinates and additional covariates in $\mathbf{X}(d)$~\citep{cressie2015statistics, gelfand2010handbook}.

While the dependence structure in equation~\ref{eq:blup} is potentially very flexible, the mean structure is quite rigid, and $\mathbf{X}(d_0)' \hat{\boldsymbol \beta}$ may be a very poor estimator of the true mean $f[\mathbf{X}(d_0)]$. In particular, assuming $f$ is linear and additive in the covariates is dubious, and the lack of regularization promotes overfitting.

In stark contrast stand a bevy of machine learning algorithms with flexible mean structures that have become popular for spatial and space-time prediction~\citep{watson2019machine,xu2018evaluation,reid2015spatiotemporal}. These models estimate $f$ with a flexible, possibly nonparametric $\tilde{f}$, but usually assume that, observations of the random field are independent of each other conditional on the covariates, i.e., that $\text{Var}[\epsilon(d_1),\epsilon(d_2)] \coloneqq 0$ for $d_1 \neq d_2$. From this it immediately follows that the expected conditional error in equation~\ref{eq:cond_exp}, $E[\epsilon(d_0) \mid \mathbf{z}(\mathbf{d})]$ must be 0, an obviously false assumption in many spatial or space-time contexts. So while $\tilde{f}[\mathbf{X}(d_0)]$ may be flexible enough to approximate $f[\mathbf{X}(d_0)]$, it is a biased estimator of $E[Y(d_0) \mid \mathbf{X}(d_0), \mathbf{z}(\mathbf{d})]$. There are certain exceptions to this characterization in the literature, and we defer a discussion of their merits to section~\ref{sec:disc}. 

The linear mean structure often assumed in kriging (and many other GP-based procedures) and the working independence assumpion often made by machine learning tools are simply not realistic. 
However, their complementary strengths suggest they might be combined into an algorithm with flexible mean \emph{and} covariance structures. We take precisely this approach, enriching regression trees with a kriging covariance model, and then ensembling them into a new prediction algorithm which we call \emph{treeging}, a portmanteau of \emph{tree} and \emph{kriging}. Treeging is implemented in a freely available R package at https://github.com/gregorywatson/treeging.  

The remainder of this manuscript proceeds as follows. In the next section, we present the treeging algorithm. In sections~\ref{sec:space} and \ref{sec:st}, we illustrate its use on spatial and space-time data respectively, including extensive simulations and case studies. Section~\ref{sec:tuning} investigates how performance depends upon tuning parameters, and section~\ref{sec:large_approx} discusses treeging in the context of large data sets. Finally, we close with a discussion.

\section{Treeging}
\label{sec:meth}

\subsection{Mean Structure} \label{subsec:mean}

We define treeging as an ensemble of base learners named \emph{treeges}, which are regression trees enriched with kriging. 
A regression tree is a heuristically defined model that recursively partitions the covariate space according to the values of covariates, modeling the conditional expectation of $Y(d)$ given $\mathbf{X}(d)$ as 
\begin{equation} \label{eq:reg_tree}
   E[Y({d})|\mathbf{X}({d})] = f[X(d), \boldsymbol \theta] = \sum_{\ell = 1}^L f_\ell [\mathbf{X}(d), \boldsymbol \theta_\ell] 1[\mathbf{X}(d) \in \mathcal{X}_\ell],
 \end{equation}
where  $\mathcal{X}_1, ...,  \mathcal{X}_L$ partition the covariate space, and  $f_\ell [\mathbf{X}(d), \boldsymbol \theta_\ell]$ gives the conditional expectation $Y(d)|\mathbf{X}(d)$ for observations with covariates that are in $\mathcal{X}_\ell$ (the $\ell$-th leave). When a constant leaf model is employed, i.e., when $f_\ell [\mathbf{X}(d), \boldsymbol \theta_\ell] = \theta_\ell$, the regression tree corresponds to a linear model with an $n~\times~L$ design matrix $\mathbf{X}_{\tau}(\mathbf{d})$ with a column for each leaf in the tree. The elements of each row of $\mathbf{X}_{\tau}(\mathbf{d})$ are 0 except for a 1 at the element corresponding to the leaf of the tree in which the covariates associated with that row reside, i.e., $[\mathbf{X}_\tau(\mathbf{d})]_{i\ell} = 1[\mathbf{X}(d_i)~\in~\mathcal{X}_\ell]$. Substituting $\mathbf{X}_\tau(d)$ for $\mathbf{X}(\mathbf{d})$ in equation~\ref{eq:blup} provides an alternative BLUP, 
\begin{equation} \label{eq:treege_blup}
\tilde{Y}(d_0) | \mathbf{X}_\tau(d_0), \mathbf{z}(\mathbf{d}) = \mathbf{X}_\tau(d_0)' \hat{\boldsymbol \beta_\tau} + \hat{\Sigma}_0 \hat{\Sigma}(\mathbf{d})^{-1} [\mathbf{y}(\mathbf{d}) - \mathbf{X}_\tau(\mathbf{d}) \hat{\boldsymbol \beta_\tau}].
\end{equation}
The superlative designation of the BLUP as \emph{best} suggests uniqueness but obscures the fact that the BLUP is specific to the choice of mean and covariance functions. Selecting an alternate mean function in equation~\ref{eq:treege_blup} thus provides an alternate BLUP. 

Like regression trees, treeges exhibit low bias and high variance in their predictions on account of their tendency to overfit. Random forest improves upon the predictive performance of a single tree by combining many in an ensemble using bootstrap samples and random subsets of the covariates to reduce the variance of the predictions~\citep{breiman2001random}. Motivated by this straightforward yet highly effective strategy, we ensembled treeges by training each one on a random subset of the training data, combining their predictions via a simple average. Differently from the iid case, data is sampled without replacement. The size of the subsample is a parameter that can be tuned, and by default we set to be $m = \floor{0.632\, n}$, because this is approximately the expected proportion of observations that appear in a bootstrap sample of size $n$~\citep{efron1997improvements}. We also mimic random forest in considering a subset of the covariates at each split. The resulting treeging prediction for $Y(d_0)$ is
\begin{equation} \label{subsampledTreeging}
  \tilde{Y}(d_0) \mid \mathbf{X}(d_0), \mathbf{z}(\mathbf{d}) = \frac{1}{T} \sum_{t=1}^T \left\{\mathbf{X}^{(t)}_{\tau}(d_0)' \hat{\boldsymbol \beta}^{(t)}_{\tau} + \hat{\Sigma}^{(t)}_0 \hat{\Sigma}\left(\mathbf{d}^{(t)}\right)^{-1} [\mathbf{y}(\mathbf{d}^{(t)}) - \mathbf{X}^{(t)}_{\tau}\left(\mathbf{d}^{(t)}\right) \hat{\boldsymbol \beta}^{(t)}_{\tau}]\right\},
\end{equation}
where $t = 1, ..., T$ indexes treeges, and a superscript $(t)$ indicates the data subset, design matrix or estimate for the $t$-th treege. 

\begin{table}
  \caption{True and estimated mean and dependence structures for spatial or space-time prediction.}
  \label{tbl:mean_and_dependence}
  \renewcommand{\arraystretch}{1.25}
  \begin{tabular}{ccc}
    \hline
      & Mean  Structure & Dependence Term \\
    \hline
Truth	&  $f[\mathbf{X}(d_0)]$ & $E[\epsilon(d_0) \mid \mathbf{z}(\mathbf{d})]$  \\
Kriging	& $\mathbf{X}(d_0)'\hat{\boldsymbol \beta}$ &  $\hat{\Sigma}_0 \hat{\Sigma}(\mathbf{d})^{-1} [\mathbf{y}(\mathbf{d}) - \mathbf{X}(\mathbf{d}) \hat{\boldsymbol \beta}]$\\
1 Regression Tree	& $\mathbf{X}_\tau(d_0)' \hat{\boldsymbol \beta}_\tau$ & 0 \\
1 Treege	& $\mathbf{X}_\tau(d_0)' \hat{\boldsymbol \beta}_\tau$ & $\hat{\Sigma}_0 \hat{\Sigma}\left(\mathbf{d}\right)^{-1} [\mathbf{y}(\mathbf{d}) - \mathbf{X}_\tau(\mathbf{d}) \hat{\boldsymbol \beta}_\tau]$ \\
Random Forest & $\frac{1}{B} \sum_{b=1}^B \mathbf{X}^{(b)}_{\tau}(d_0)' \hat{\boldsymbol \beta}^{(b)}_{\tau}$& 0   \\
Treeging	& $\frac{1}{T} \sum_{t=1}^T \mathbf{X}^{(t)}_{\tau}(d_0)' \hat{\boldsymbol \beta}^{(t)}_{\tau}$ & $\frac{1}{T} \sum_{t=1}^T \hat{\Sigma}^{(t)}_0 \hat{\Sigma}\left(\mathbf{d}^{(t)}\right)^{-1} [\mathbf{y}(\mathbf{d}^{(t)}) - \mathbf{X}^{(t)}_{\tau}\left(\mathbf{d}^{(t)}\right) \hat{\boldsymbol \beta}^{(t)}_{\tau}]$ \\
    \hline
  \end{tabular}
\end{table}

Table~\ref{tbl:mean_and_dependence} lists the mean and dependence structures for kriging, random forest, a single treege and treeging. The limitations of kriging and random forest are apparent as is the manner in which they have been combined into treeging. 

\subsection{Covariance Estimation} \label{subsec:cov}

Because a treege uses the same covariance structure as kriging, it accommodates any covariance estimation procedure that works for kriging. This makes available the wide variety of covariance structures developed for kriging, including procedures that iteratively estimate the mean and covariance, and importantly allows for the incorporation, of domain specific knowledge~\citep{stein2005space,lloyd2010local,boisvert2009kriging}. It is also possible to use different covariance function estimators for different treeges within a treeging ensemble. The default covariance estimation strategy used for the results in this manuscript employs spherical covariance functions. Briefly, for each treege we fit a spherical variogram function to the empirical variogram of the regression tree residuals to estimate the parameters of a spherical covariance function. For space-time data, we fit separable spherical covariance functions to the spatial and temporal distances. The details of these procedures may be found in the appendix. One convenient feature of the spherical covariance function is that covariance decreases to 0 at a finite distance unlike other common choices, such as the exponential function. This can be exploited for computational expediency as we discuss in section~\ref{sec:large_approx}. 

\subsection{Treeging Algorithm}

Algorithm~\ref{alg:treeging} describes the treeging algorithm in pseudocode for predicting a vector of outcomes, $\mathbf{y}(\mathbf{d}_0)$. For each of the $T$ treeges, a random subsample of size  
$\floor{n\,p}$, 
$p\in(0,1)$, 
of the observed data is used to train a regression tree. Using an optimization procedure, the parameters of a spherical semi-variogram, $\boldsymbol\varphi$ are fit to the residuals of the regression tree,  $\mathbf{e}^{(t)}_\tau\left(\mathbf{d}^{(t)}\right)$, and their pairwise distances, $h_{ij} = \lVert d_i^{(t)} - d_j^{(t)} \rVert$, for $i,j = \{1,..., \floor{np}\}$. The estimated spherical parameters $\boldsymbol \varphi^{(t)}$ provide an estimate of the covariance function, $\hat{\Sigma}^{(t)}$. Using this covariance function and the regression tree design matrix $\mathbf{X}^{(t)}(\mathbf{d})$, the treege prediction can be computed according to equation~\ref{eq:treege_blup}. Averaging the predictions of the $T$ treeges yields the treeging prediction, $\tilde{\mathbf{y}}(\mathbf{d}_0)$. 

\begin{algorithm}
\caption{Treeging}\label{alg:treeging}
\begin{algorithmic}[1]
\Procedure{Treeging}{$\mathbf{z}(\mathbf{d}), \mathbf{d}_0, \mathbf{X}_0(\mathbf{d}_0), p, T$}
\For {$t \gets 1 \text{ to } T$}
  \State $\mathbf{z}\left(\mathbf{d}^{(t)}\right) \gets \text{A random subset of size  $\floor{np}$  from } \mathbf{z}(\mathbf{d})$
  \State  $\left\{\mathbf{X}^{(t)}_\tau(d),  \hat{\boldsymbol \beta}^{(t)}_{\tau}\right\} \gets \text{Train a regression tree on }  \mathbf{z}\left(\mathbf{d}^{(t)}\right)$ 
  \State  $\mathbf{e}^{(t)}_\tau\left(\mathbf{d}^{(t)}\right) \gets \mathbf{y}\left(\mathbf{d}^{(t)}\right) - \mathbf{X}^{(t)}_\tau\left(\mathbf{d}^{(t)}\right)' \hat{\boldsymbol \beta}^{(t)}_{\tau}$ 
    \State $\gamma^{(t)}\left(h; \boldsymbol \varphi^{(t)}\right) \gets$ Fit a spherical semi-variogram to $\mathbf{e}^{(t)}_\tau\left(\mathbf{d}^{(t)}\right)$ and their pairwise distances
  \State $\hat{\Sigma}^{(t)}_{ij} \gets {\Sigma}\left(d^{(t)}_i, d^{(t)}_j;\boldsymbol\varphi^{(t)} \right), i,j = 1, ..., \text{round}(np)$
   \State $\hat{\Sigma}^{(t)}_0 \gets {\Sigma}\left(\mathbf{d}_0, \mathbf{d}^{(t)};\boldsymbol\varphi^{(t)} \right)$   
  \State $\tilde{\mathbf{y}}^{(t)}(\mathbf{d}_0) \gets \mathbf{X}^{(t)}_{\tau}(d_0) \hat{\boldsymbol \beta}^{(t)}_{\tau} + \hat{\Sigma}^{(t)}_0 {\hat{\Sigma}^{(t)}}\strut^{-1} \left[\mathbf{y}\left(\mathbf{d}^{(t)}\right) - \mathbf{X}^{(t)}_{\tau}\left(\mathbf{d}^{(t)}\right) \hat{\boldsymbol \beta}^{(t)}_{\tau}\right]$
\EndFor
\State $\tilde{\mathbf{y}}(\mathbf{d}_0) \gets \frac{1}{T}\sum_{t=1}^{T} \tilde{\mathbf{y}}^{(t)}(\mathbf{d}_0)$
\EndProcedure
\end{algorithmic}
\end{algorithm}


\section{Spatial Treeging}
\label{sec:space}

We investigated the suitability of treeging for spatial interpolation by comparing it to random forest, kriging and a kriging ensemble in a series of simulations and two case studies. The kriging ensemble is constructed using the same subsampling and ensembling procedures as treeging, but with a linear mean structure for the base learners rather than the tree-based mean structure of treeges. Kriging ensembles are not commonly used, however, we included it in the comparison to discern the predictive gains or losses associated with treeging's flexible mean structure from those of ensembling. 

We used $R^2$ as our primary measure of predictive accuracy, which is the ratio of the mean squared prediction error to the variance, 
\begin{equation}
   R^2 (\mathbf{y}, \hat{\mathbf{y}}) =1 - 
    \frac{\sum_{i=1}^n (y_i - \hat{y}_i)^2}{\sum_{i=1}^n (y_i - \bar{y})^2},
\end{equation}
where $\bar{y}$ is the mean of $\mathbf{y}$. This is often interpreted as the proportion of the variation in $\mathbf{y}$ explained by predicted values $\hat{\mathbf{y}}$, but the $R^2$ can be negative if it is a worse predictor than $\bar{y}$, which is quite possible in out-of-sample prediction in which case the sample mean is unknown. In the simulation setting, the true $R^2$ can be computed. In the case studies, we estimate the $R^2$ using 10-fold cross-validation~(CV), denoting the estimate as $\hat{R}^2_{CV}$. 

\subsection{Spatial Simulation Study}
\label{subsec:space_sim}

We simulated 1,029 spatial random fields across a variety of covariate effect sizes and levels of spatial dependence. In each case, the field was generated from a Gaussian process with a mean function depending on 3 covariates, one with a linear effect, one a threshold effect, and one a sigmoidal effect to mimic  what is often encountered in environmental pollution studies. The mean function was scaled up or down using an effect size multiplier to elucidate the relationship between predictive performance and covariate effect size.  
The appendix provides the mean function and simulation parameters. We randomly selected 100 locations in a spatial domain as the observed data on which to train the models, and assessed their predictive accuracy over a $21~\times~21$ grid of points across the spatial field under 4 different covariate scenarios: (i)~the 3~informative covariates, (ii)~the 3~informative covariates plus 17~spurious covariates, (iii)~2 of the 3 informative covariates selected at random, and (iv)~2 of the 3 informative covariates plus the 17 spurious covariates. In addition, the spatial coordinates were included as covariates in all cases. These 4 scenarios were constructed to reflect some of the common challenges encountered in spatially dependent data. Often there are potentially informative ancillary data, but which covariates are informative is unknown a~priori, and not all relevant covariates may be available. 

\begin{figure}
  \begin{center}
  \begin{tabular}{c}
   A. Predictive Accuracy as a Function of Spatial Dependence \\ 
   \includegraphics[width=6.8in]{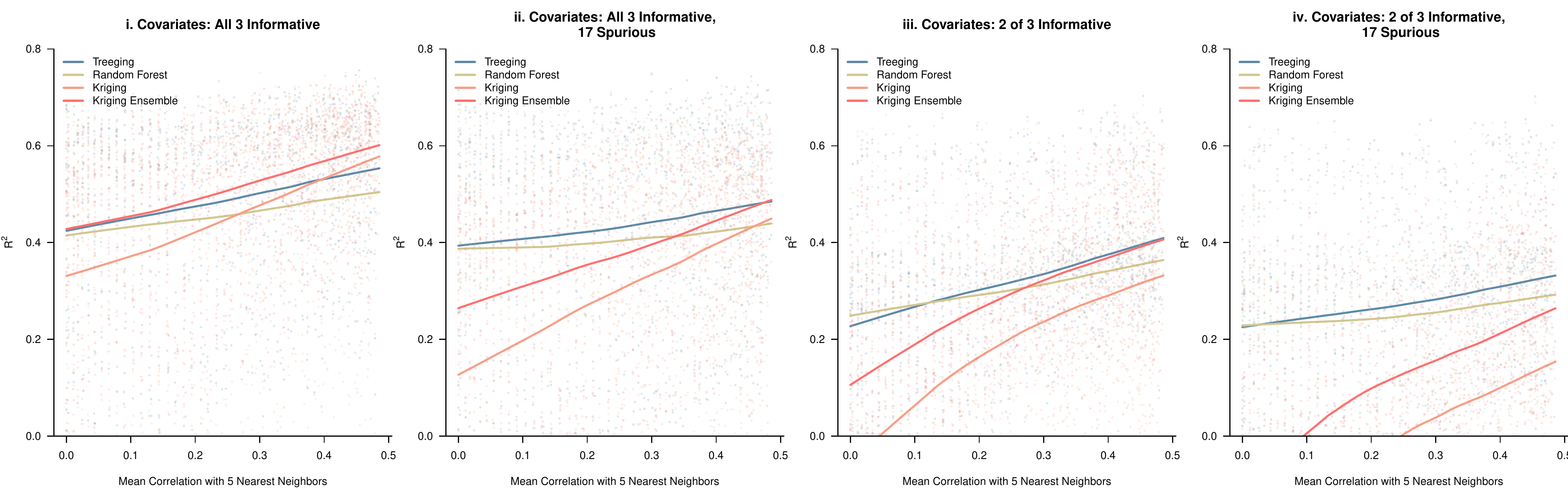} \\ 
   B. Predictive Accuracy as a Function of Covariate Effect Size \\   
   \includegraphics[width=6.8in]{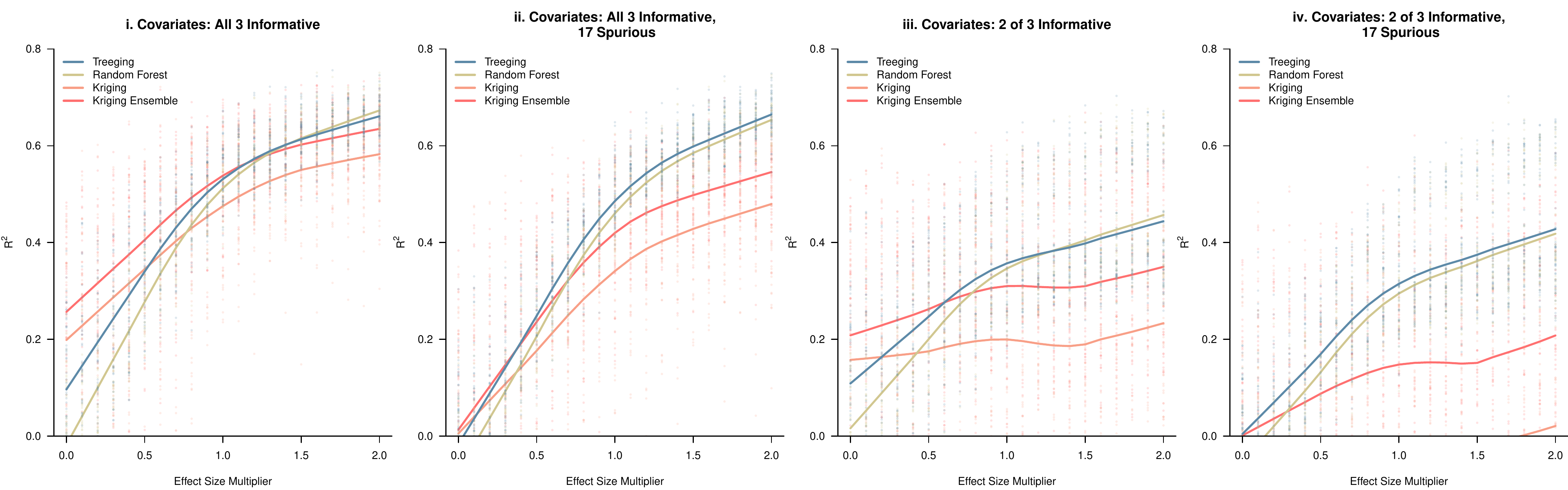} \\  
   C. Most Accurate Model as a Function of Spatial Dependence and Covariate Effect Size \\   \includegraphics[width=6.8in]{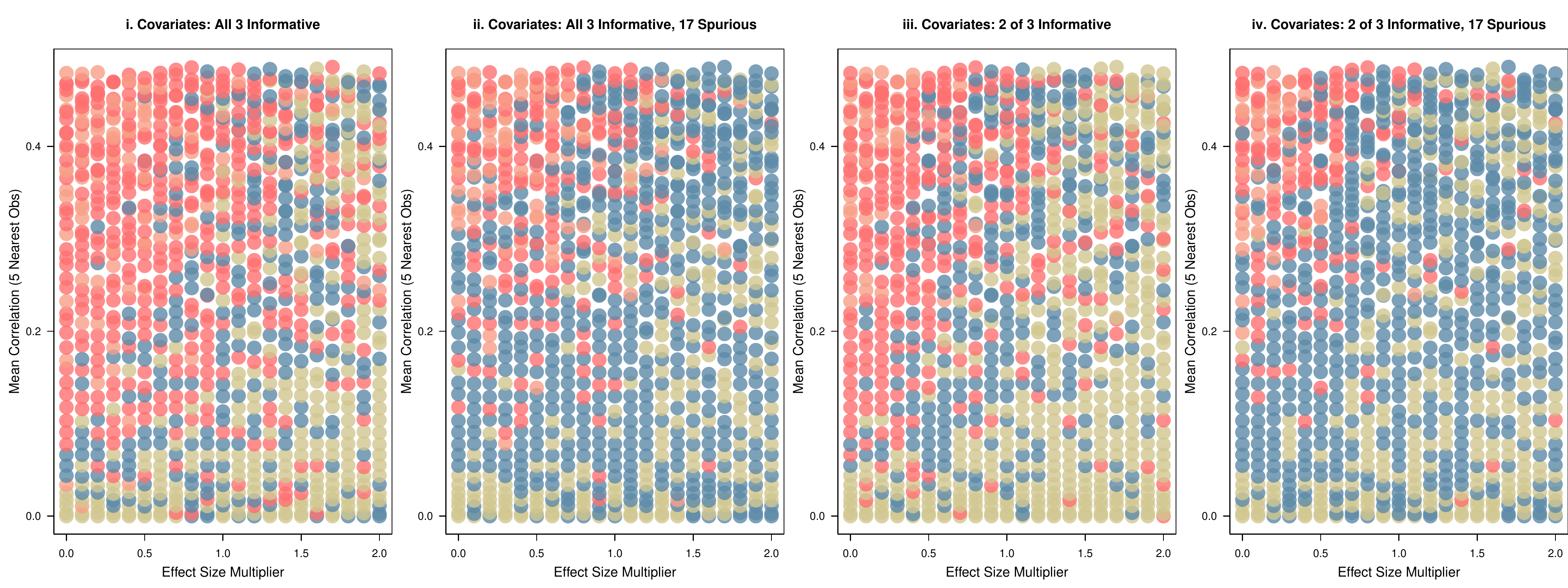}   

  \end{tabular}
  \end{center}
  \caption{Spatial Simulation Model Comparison: Predictive accuracy ($R^2$) as a function of (A.) spatial dependence, (B.)  covariate effect size, and (C.) both. In (C.) the color of each dot depicts the  highest $R^2$ model for the simulation at that combination of the effect size multiplier and mean correlation.}
   \label{fig:space_sim_model_comp}
\end{figure}

Figure~\ref{fig:space_sim_model_comp}A depicts the predictive $R^2$ for each model under each covariate scenario as a function of the spatial dependence, which is assessed as the mean correlation between the prediction locations and the five nearest training data points. The results are smoothed using a LOWESS curve to facilitate comparison. Not surprisingly, there is a general increasing trend in $R^2$ as spatial dependence increases for all models across every scenario, as increased spatial dependence corresponds to additional information with which to predict $Y(d)$ over the unobserved grid. This increase is less pronounced for random forest, which ignores dependence, although it is able to make some gains, because the spatial coordinates are included as covariates. 

The kriging ensemble is the best performing model when trained on all three informative coviarates without any spurious covariates (scenario~i), with treeging second best until it is overtaken by kriging at high levels of dependence. In the presence of spurious coviarates (scenario~ii), treeging is the best model with the kriging ensemble performing comparably at high levels of dependence. In scenarios~iii and~iv, kriging and the kriging ensemble perform very poorly. Across all four scenarios, random forest performs respectably, but treeging outpferforms it across the board with a small margin between them at low spatial correlation that increases as dependence gets stronger.

Figure~\ref{fig:space_sim_model_comp}B plots the marginal effect size multiplier against $R^2$ for the same set of simulations. Across all four covariate scenarios there is again a consistent relationship between treeging and random forest. When the effect size multiplier is 0, the covariate effects are 0 and random forest performs poorly. Treeging performs noticeably better, although it is also not particularly accurate. As covariate effect strengths increase, the $R^2$ for both treeging and random forest increases rapidly with random forest gradually closing the gap between them. Kriging and ensembled kriging perform fairly well when trained on all the informative covariates (scenarios~i and~ii) with the ensemble generally superior. At 0 effect size they have higher $R^2$ than treeging and random forest, but are overtaken as effect size increases. In scenarios~iii and~iv, however, they perform very poorly with $R^2$ decreasing as covariate effect size increases. Despite this additional predictive information, the mean structure of kriging is apparently too rigid when an informative covariate is withheld. 

Figure~\ref{fig:space_sim_model_comp}C plots effect size against mean correlation with the color of each dot depicting the best (i.e., highest $R^2$) model for that simulation. At low effect sizes (the left edge), kriging and ensembled kriging are generally best except when there is very little or no spatial dependence. When there is low dependence (the bottom edge), random forest and treeging are best. With the exception of scenario~i, when there is substantial spatial dependence and effect size, treeging tends to be best. 

\subsection{Spatial Case Studies}
\label{subsec:space_case}

We compared the predictive accuracy of the same set of models on daily 8-hour maximum average ozone and 24-hour average fine particulate matter (PM$_{2.5}$) concentrations during wildfires in California in 2008~\citep{watson2019machine}. Data were collected from 100 ozone and 45 PM$_2.5$ monitors over 118 and 120 days respectively. In addition to these data, 15 covariates including atmospheric weather data, land use information and chemical transport model predictions were collected along with the spatial coordinates. These covariates and their sources are listed in the appendix. To compare their accuracy at strictly spatial interpolation, we considered each day of the data as a distinct data set and estimated predictive accuracy using 10-fold~CV, taking the average of each day's 10 folds as the estimated $\hat{R}^2_{CV}$ for that day. In Section~\ref{subsec:st_case} we evaluate interpolation on the entire data set in our discussion of space-time treeging, which is a more realistic approach to interpolation on these data.  

\begin{figure}
\includegraphics[width=3.4in]{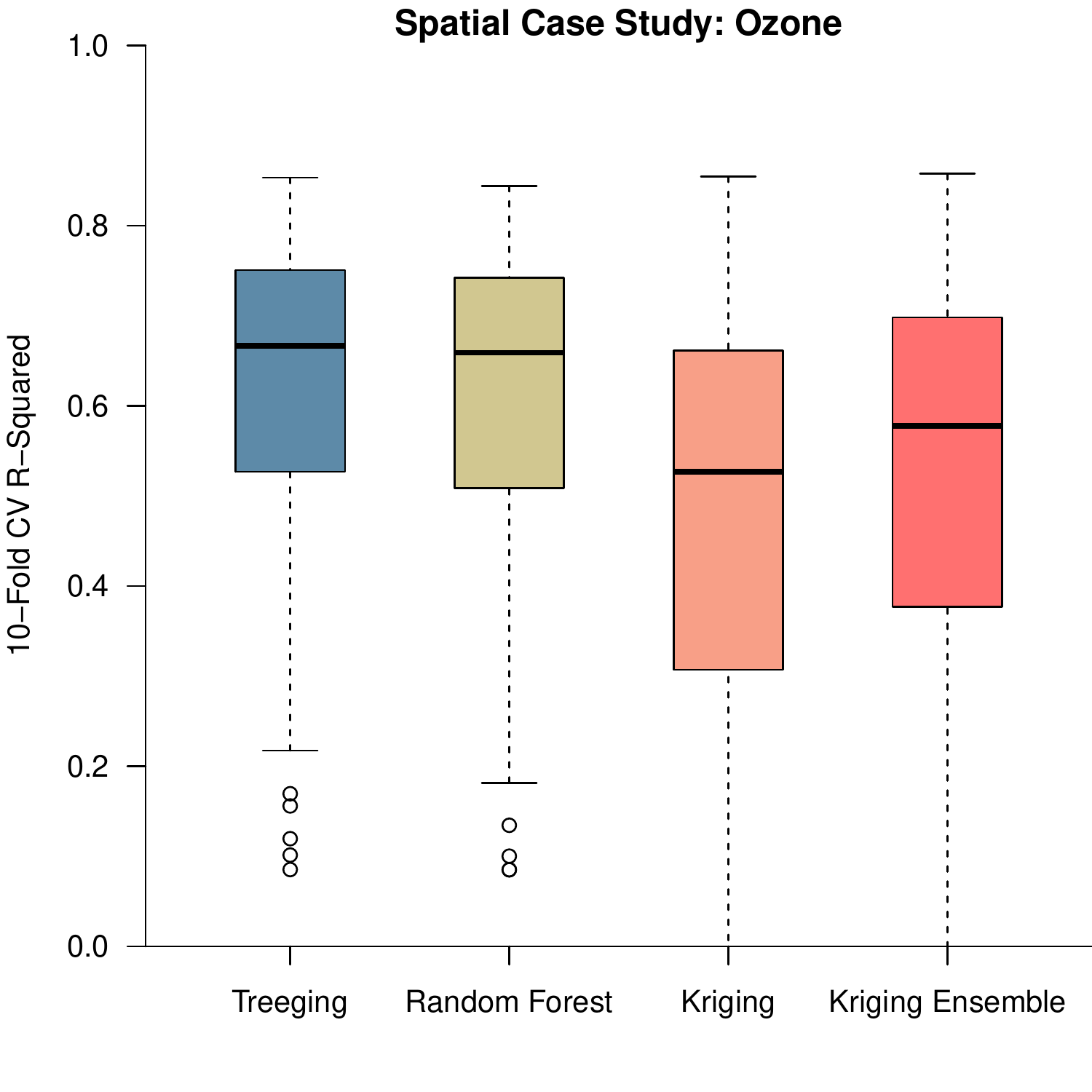}
\includegraphics[width=3.4in]{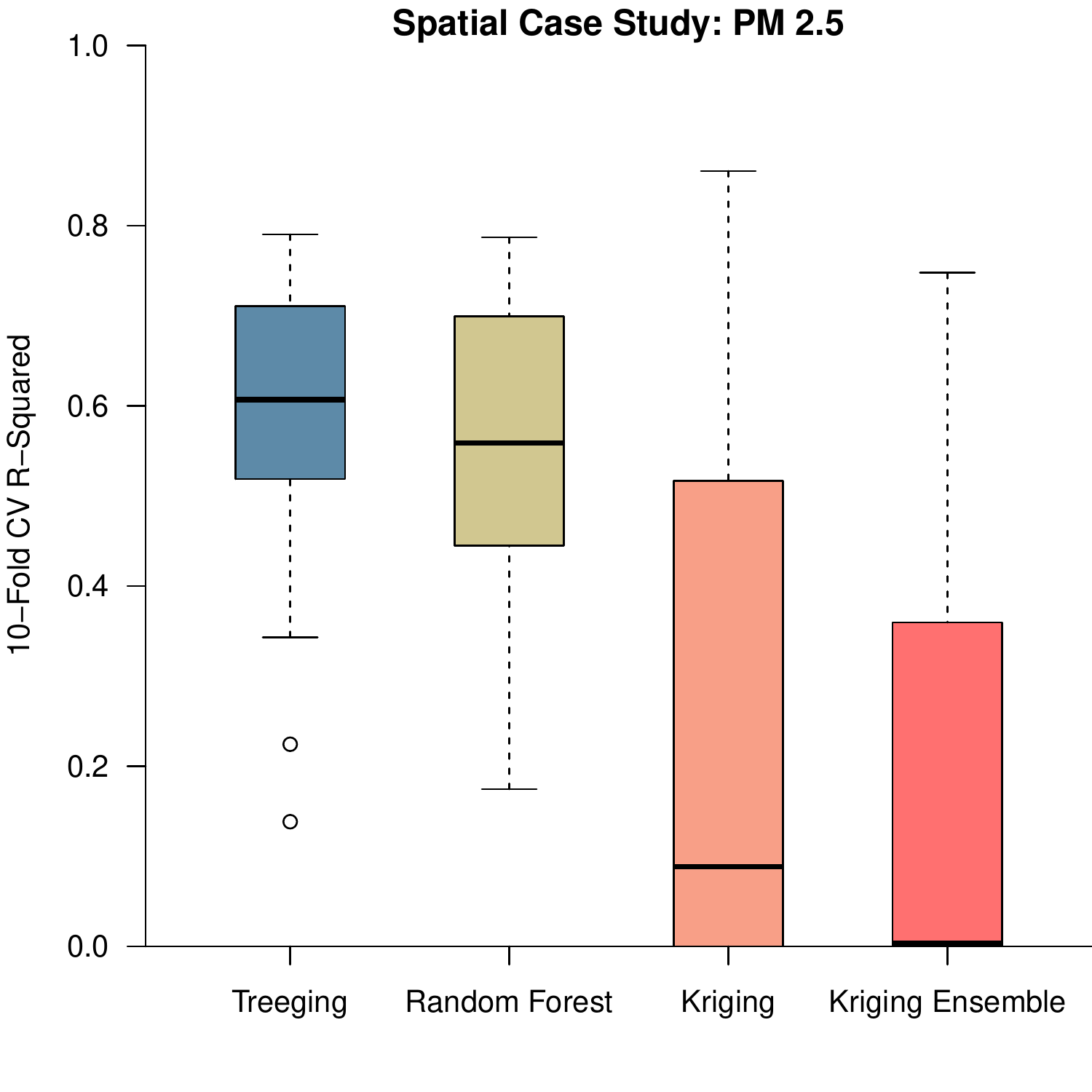}
  \caption{Spatial Case Studies Model Comparison: 10-fold~CV estimates of $R^2$ were produced for each day, treating each day as an independent spatial data set. The resulting distribution of estimates is summarized for each model as a boxplot, with the darkened central line indicating the median $\hat{R}_{CV}^2$ across all days for that model.}
  \label{fig:space_case_model_comp}
\end{figure}

Figure~\ref{fig:space_case_model_comp} depicts the average $\hat{R}^2_{CV}$ on each day for each of the models on PM$_{2.5}$ and ozone as box and whisker plots. For both pollutants, treeging was the most accurate model (ozone $\hat{R}^2_{CV}$:~0.62, PM$_{2.5}$ $\hat{R}^2_{CV}$:~0.58), performing slightly better than random forest (ozone $\hat{R}^2_{CV}$:~0.61, PM$_{2.5}$ $\hat{R}^2_{CV}$:~0.56). For ozone, the kriging models are somewhat less accurate on average (kriging $\hat{R}^2_{CV}$:~0.43, ensemble $\hat{R}^2_{CV}$:~0.48), and worryingly have substantially higher variance. For PM$_{2.5}$, this variation is even greater, and the CV $\hat{R}^2_{CV}$ is also much lower (kriging $\hat{R}^2_{CV}$:~0.00, ensemble $\hat{R}^2_{CV}$:~-0.24). These results suggest that kriging models may be less stable than models with built-in feature selection like treeging and random forest.

\section{Space-Time Treeging}
\label{sec:st}

Space-time treeging considers prediction of a random field $Y(d)$ indexed over a 3-dimensional space-time domain, $\mathcal{D} \subset \mathcal{R}^3$. In the context of space-time data, multiple types of prediction are possible~\citep{watson2020prediction}. We consider here the suitability of treeging for space-time interpolation, a common type of space-time prediction in which repeated measurements are taken at a set of locations and the predictive goal is to predict the entire time series at new spatial locations. This specific predictive target is motivated by typical applications in environmental pollution studies, which require the estimation of pollutant concentration as a potential source of exposure at unobserved spatial locations~\citep{watson2020prediction}. 

\subsection{Space-Time Simulation Study}
\label{subsec:st_sim}

We conducted a second battery of simulations to investigate the predictive accuracy of space-time treeging. The random field $Y(d)$ was sampled as a Guassian process over a time series of length 30 at each of 40 training locations and an $11~\times~11$ grid of test locations. The spatial dependence, temporal dependence and covariate effect sizes were varied to investigate their impact on predictive accuracy. The precise details and the functional forms of the covariates may be found in the appendix. Similarly to the spatial simulation study described above, we considered four covariate scenarios. 

\begin{figure}
  \begin{center}
  \begin{tabular}{ c}
   A. Predictive Accuracy as a Function of Spatial Dependence \\ 
   \includegraphics[width=6.8in]{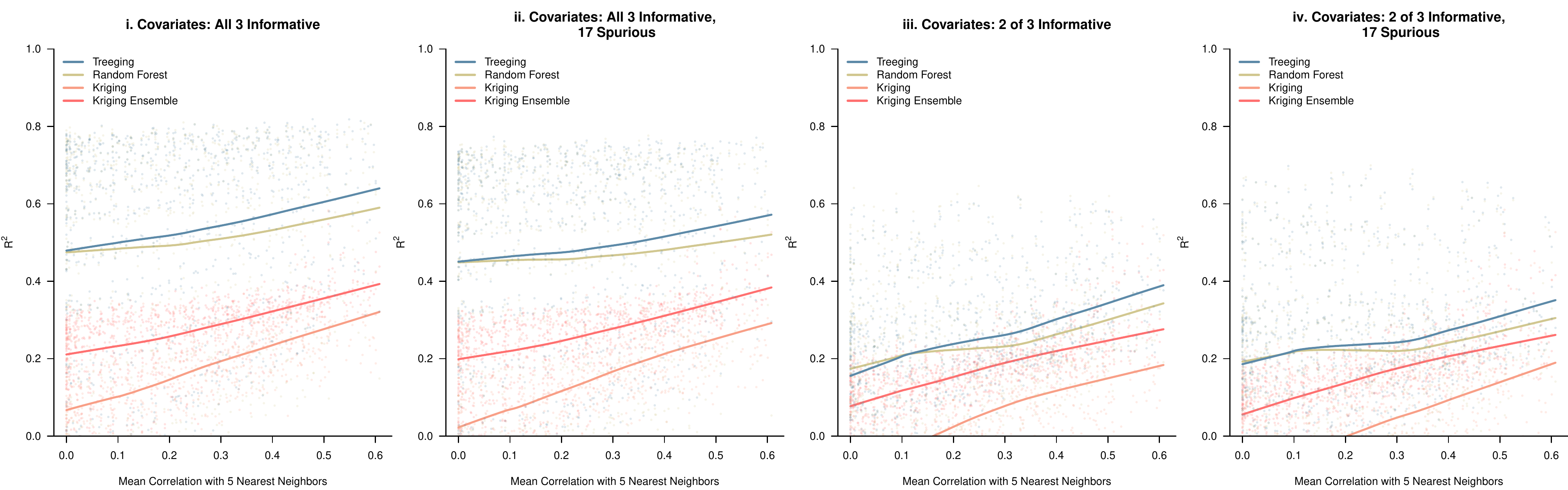} \\ 
   B. Predictive Accuracy as a Function of Covariate Effect Size \\   
   \includegraphics[width=6.8in]{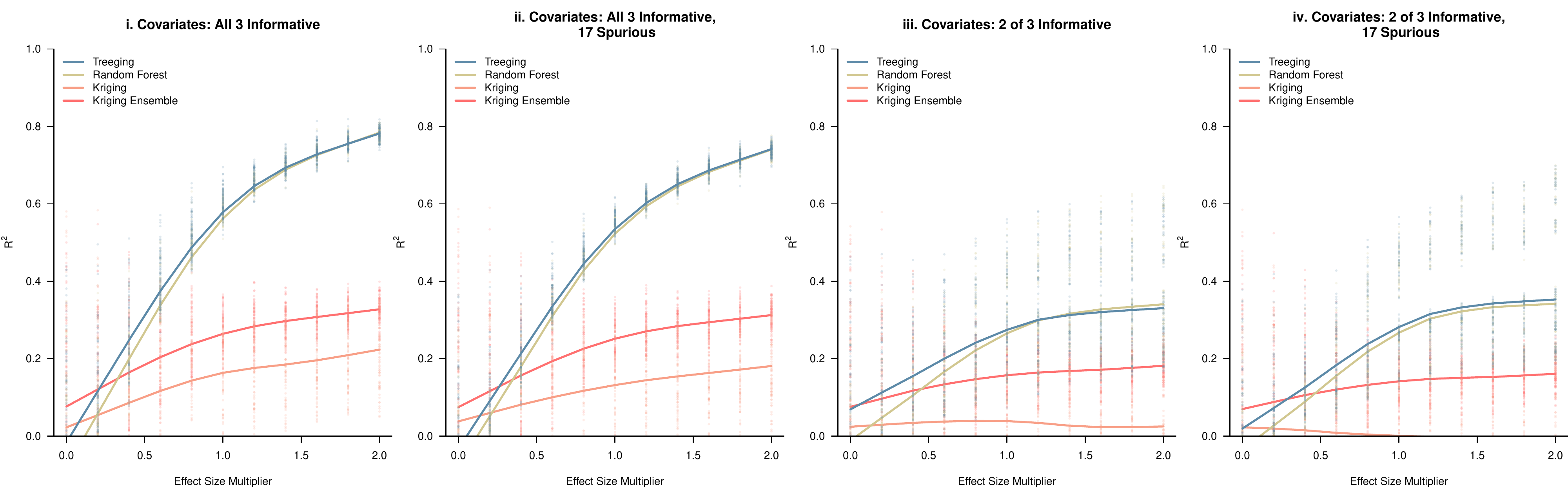} \\  
   C. Most Accurate Model as a Function of Spatial Dependence and Covariate Effect Size \\   \includegraphics[width=6.8in]{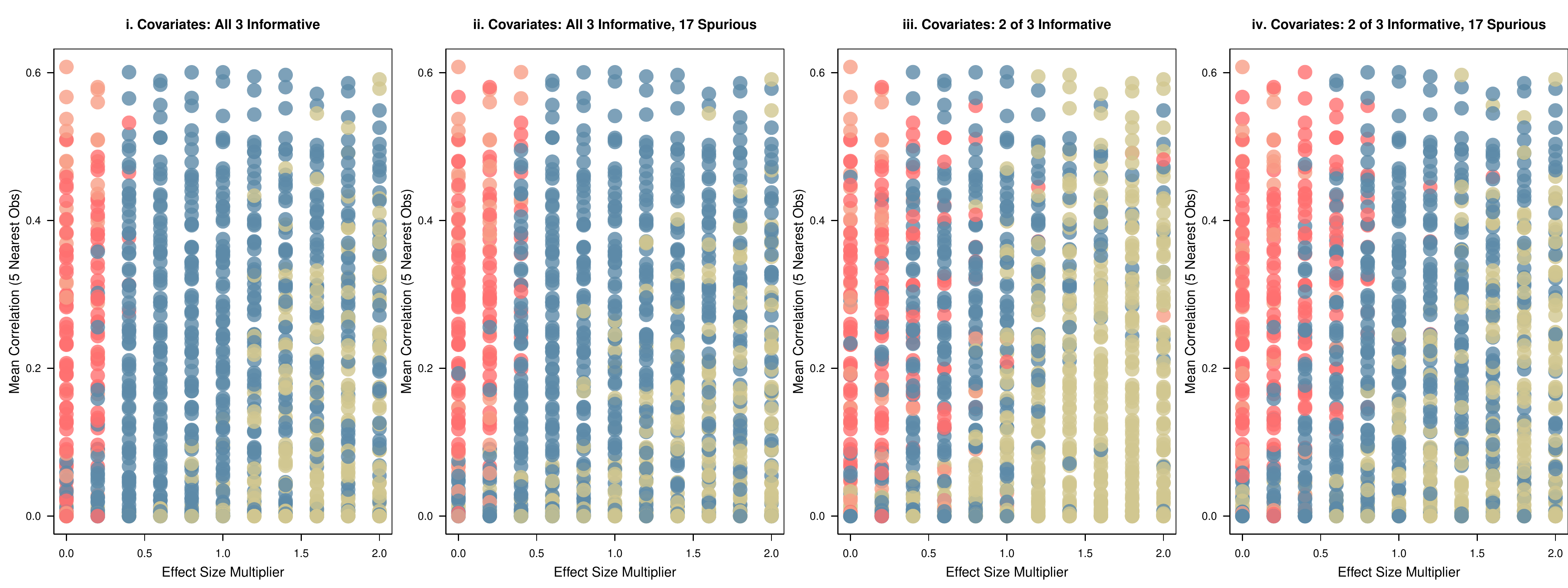}    

  \end{tabular}
  \end{center}
  \caption{Space-Time Simulation Model Comparison: Predictive accuracy ($R^2$) as a function of (A.) spatial dependence, (B.)  covariate effect size, and (C.) both. In (C.) the color of each dot depicts the  highest $R^2$ model for the simulation at that combination of the effect size multiplier and mean correlation.}
   \label{fig:st_sim_model_comp}
\end{figure}

Figure~\ref{fig:st_sim_model_comp} shows the results of the simulation study. Figure~\ref{fig:st_sim_model_comp}A depicts predictive $R^2$ as a function of the mean correlation between test points and the five training points with which they are most highly correlated. The functional relationship between mean correlation and $R^2$ is estimated using a LOWESS curve for each of the four models. In all four of the covariate scenarios, treeging (blue) and random forest (green) have nearly identical $R^2$ when the mean correlation is~0, i.e., when there is no dependence, which is the left hand extreme of the plots. As the mean correlation increases, the $R^2$ for random forest increases slightly, while that of treeging improves more rapidly, resulting in increasing divergence between these two curves. Treeging better exploits the additional information in the increasing dependence between training and test points, because it expliticly models this dependence, while random forest assumes there is none. Kriging (orange) and ensembled kriging (red) also improve as the mean correlation increases, since they too model the space-time dependence. Kriging is consistently below the kriging ensemble, which is not surprising given the well known improvements in predictive accuracy associated with ensembling base learners. 
 
Figure~\ref{fig:st_sim_model_comp}B depicts predictive $R^2$ as a function of covariate effect size for the same set of simulations. Again kriging lags the kriging ensemble by a consistent amount. When the effect size multiplier is 0, the covariates have no effect on the outcome, aside from possible marginal associations with space or time resulting from the space-time dependence, random forest has the worst $R^2$ in all four scenarios, because it relies exclusively on the relationship between the covariates and the outcome. Not surprisingly it is at this extreme that treeging excels random forest by the largest margin. It is also at this extreme that the kriging models tend to predict better than treeging. It would seem that treeging's flexible mean structure hurts its accuracy in the absence of informative covariates, likely on account of the greater potential for overfitting. As the effect size multiplier increases, the covariates exert a stronger effect, treeging and random forest surpass the kriging models on the strength of their flexible mean structures, and random forest gradually closes the gap with treeging as the information in the covariates dominates that of the space-time dependence. 

Figure~\ref{fig:st_sim_model_comp}C combines the marginal plots of figures~\ref{fig:st_sim_model_comp}A and~\ref{fig:st_sim_model_comp}B, plotting effect size against mean correlation with each simulation as a single dot. The color of the dot indicates which of the four models had the highest $R^2$ for that simulation. While this depiction ignores the margin of victory by which a model is best, it nonetheless offers a useful graphical summary of model performance. In all four covariate scenarios, the kriging models tend to perform best at very low effect sizes (the left edge of the plots). Random forest, and to a lesser extent, treeging tend to be best at very low mean correlation (the bottom edge of the plots). In between, treeging tends to dominate across a wide swath of the plots. This nicely illustrates the benefits of treeging---while it may not be the best model in all cases, it is the generally the best when both the mean structure and dependence structures are informative and competitive when only one is.   

\subsection{Space-Time Case Studies}
\label{subsec:st_case}

We evaluated the space-time predictive performance of treeging using the same PM$_{2.5}$ and ozone data used in section~\ref{subsec:space_case} for spatial case studies. To provide insight into the potential impact of time series length, we estimated predictive accuracy on increasingly long subsets of the data from one to eight weeks. We used 10-fold location-based CV to estimate predictive accuracy in which the monitor locations were partitioned into 10 folds, so that all observations taken at a particular spatial location fall in the same fold. This is critical to avoid overly optimistic estimates of spatial interpolation prediction error~\citep{watson2020prediction}. 

\begin{figure}
\includegraphics[width=3.4in]{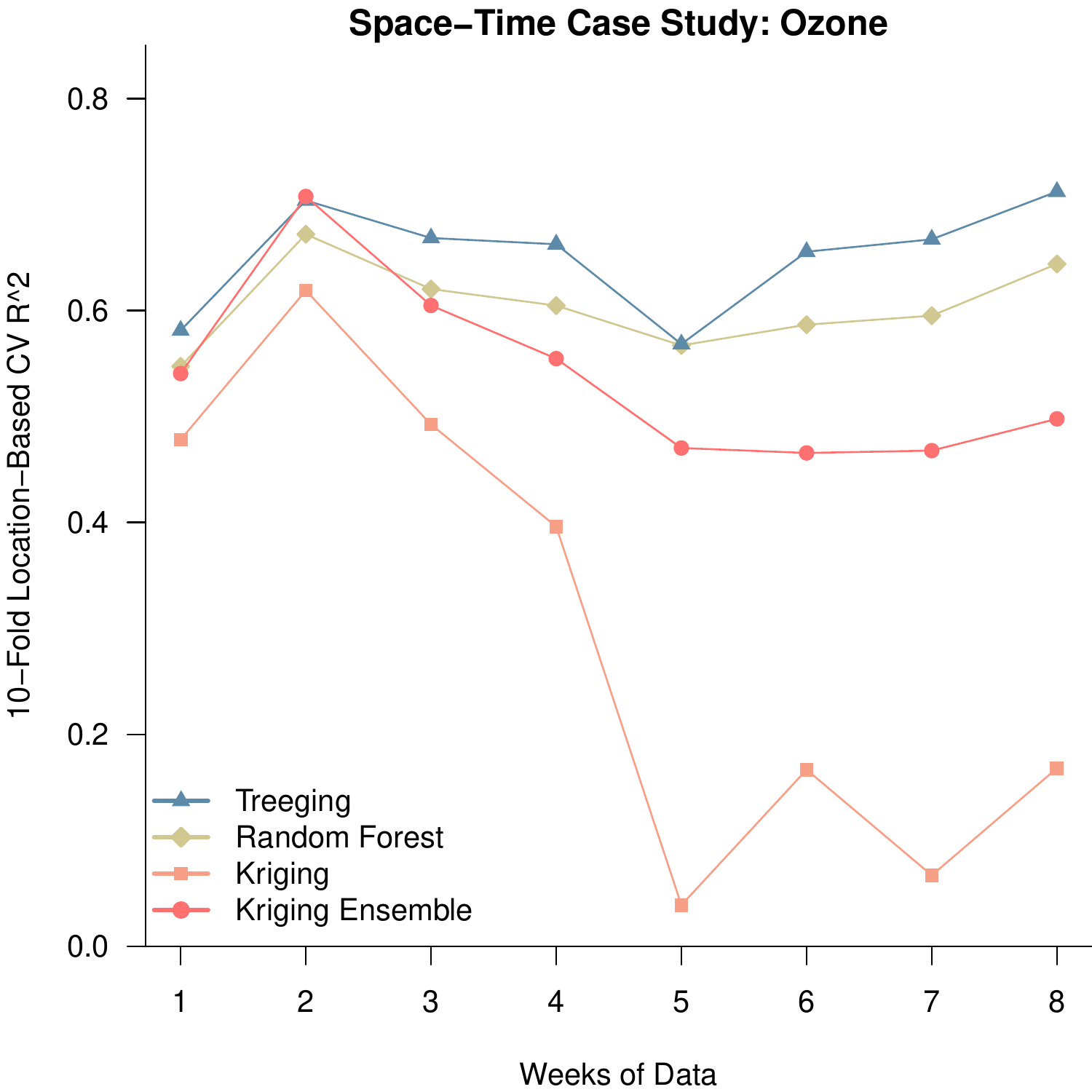}
\includegraphics[width=3.4in]{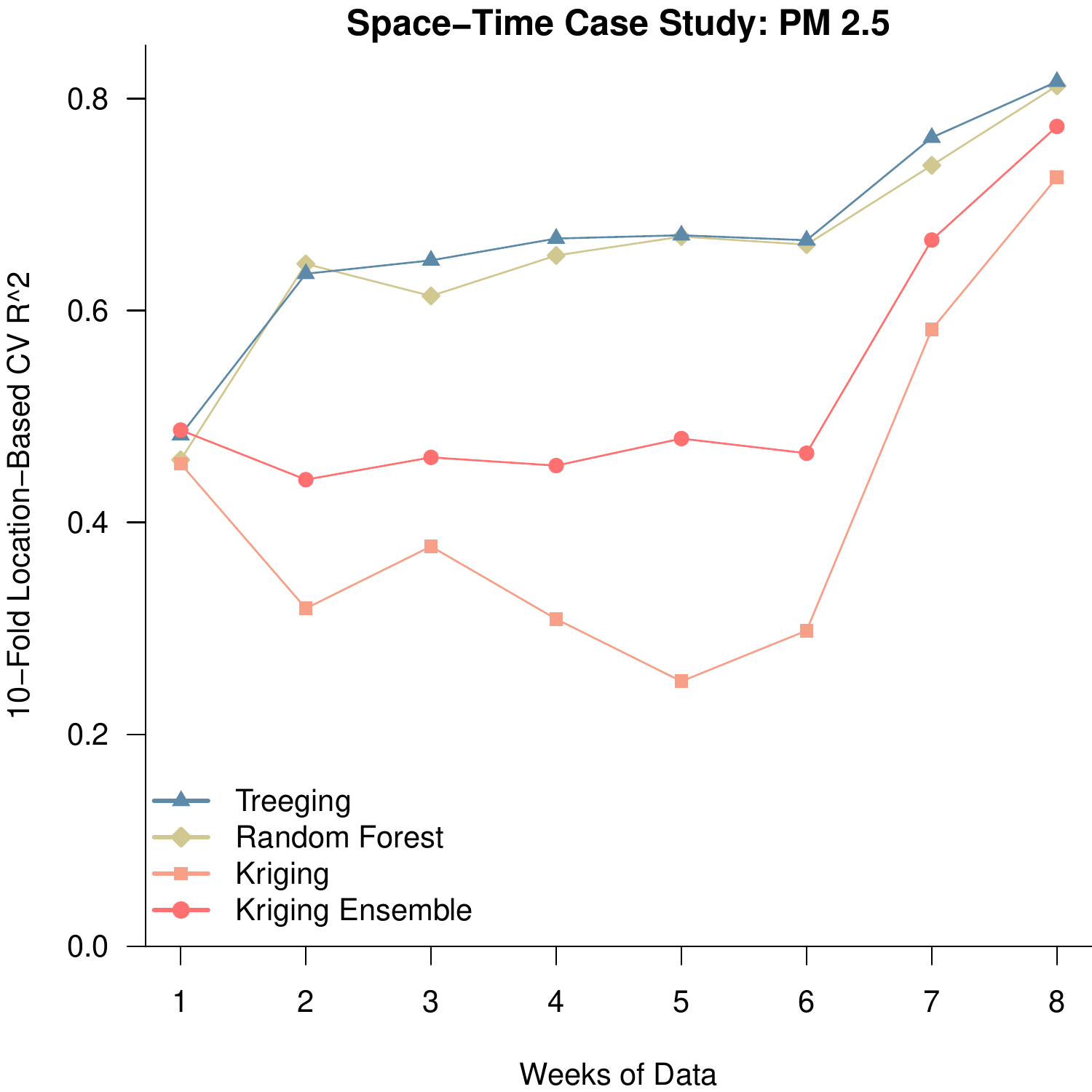}
  \caption{Space-Time Case Studies Model Comparison. }
  \label{fig:st_case_model_comp}
\end{figure}

Figure~\ref{fig:st_case_model_comp} graphically depicts the $\hat{R}^2_{CV}$ case study results. For both PM$_{2.5}$ and ozone, treeging is the best performing model, followed by random forest then ensembled kriging and lastly kriging. For PM$_{2.5}$, the difference in $\hat{R}^2_{CV}$ between treeging and random forest is very slight. For ozone, however, treeging does provide a substantial improvement over random forest. This suggests that the covariates may be more informative for interpolating PM$_{2.5}$, similar to simulation scenario~iii, while there may be informative covariates for ozone that are not contained in this data, as in simulation scenario~iv.  

The $\hat{R}^2_{CV}$ for kriging and to a lesser extent ensembled kriging are both lower and more highly variable than treeging and random forest in figure~\ref{fig:st_case_model_comp}. The accuracy of kriging could potentially be improved with the use of a more sophisticated covariance estimation procedure. This may benefit kriging relative to random forest, but a more accurate covariance model would also benefit treeging, which can use any covariance structure that kriging can. The greater variability of kriging estimates reflects the potential for overfitting with a rigid mean structure that lacks a mechanism for feature selection. This is mitigated to some extent by the model averaging in the ensemble, but there is still greater variability than the tree-based ensembles.

\section{Parameter Tuning}
\label{sec:tuning}

\begin{figure}
\includegraphics[width=2.3in]{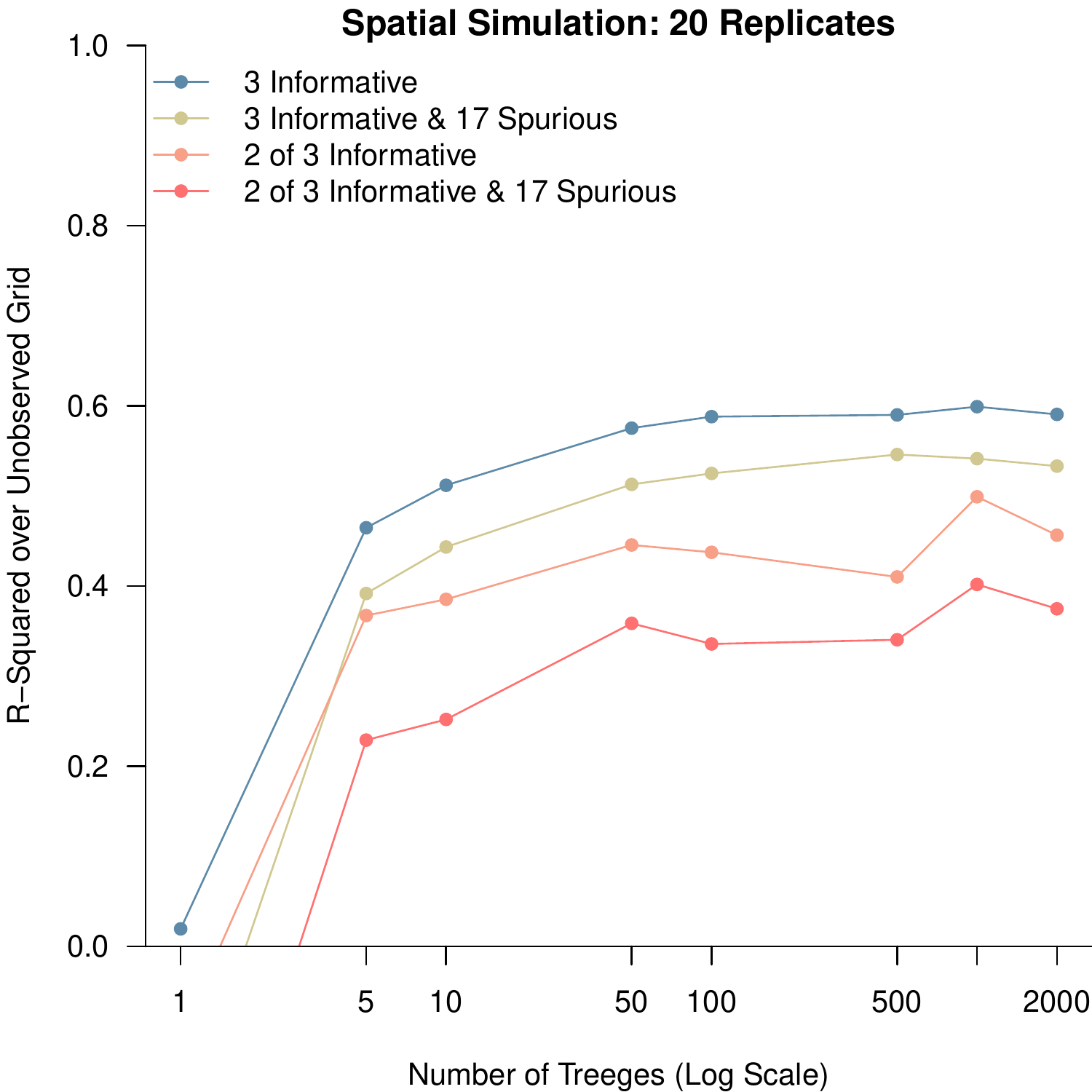}
\includegraphics[width=2.3in]{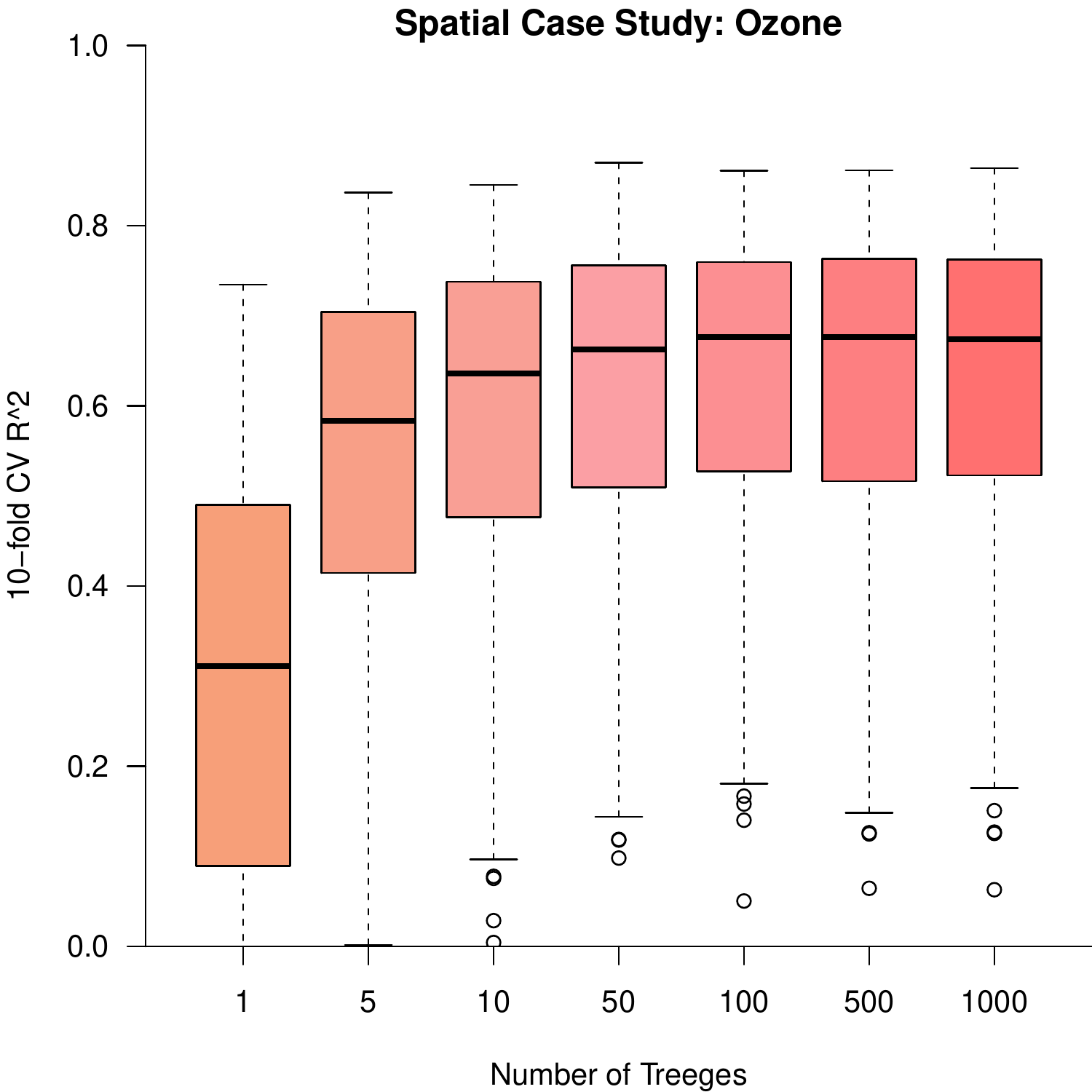}
\includegraphics[width=2.3in]{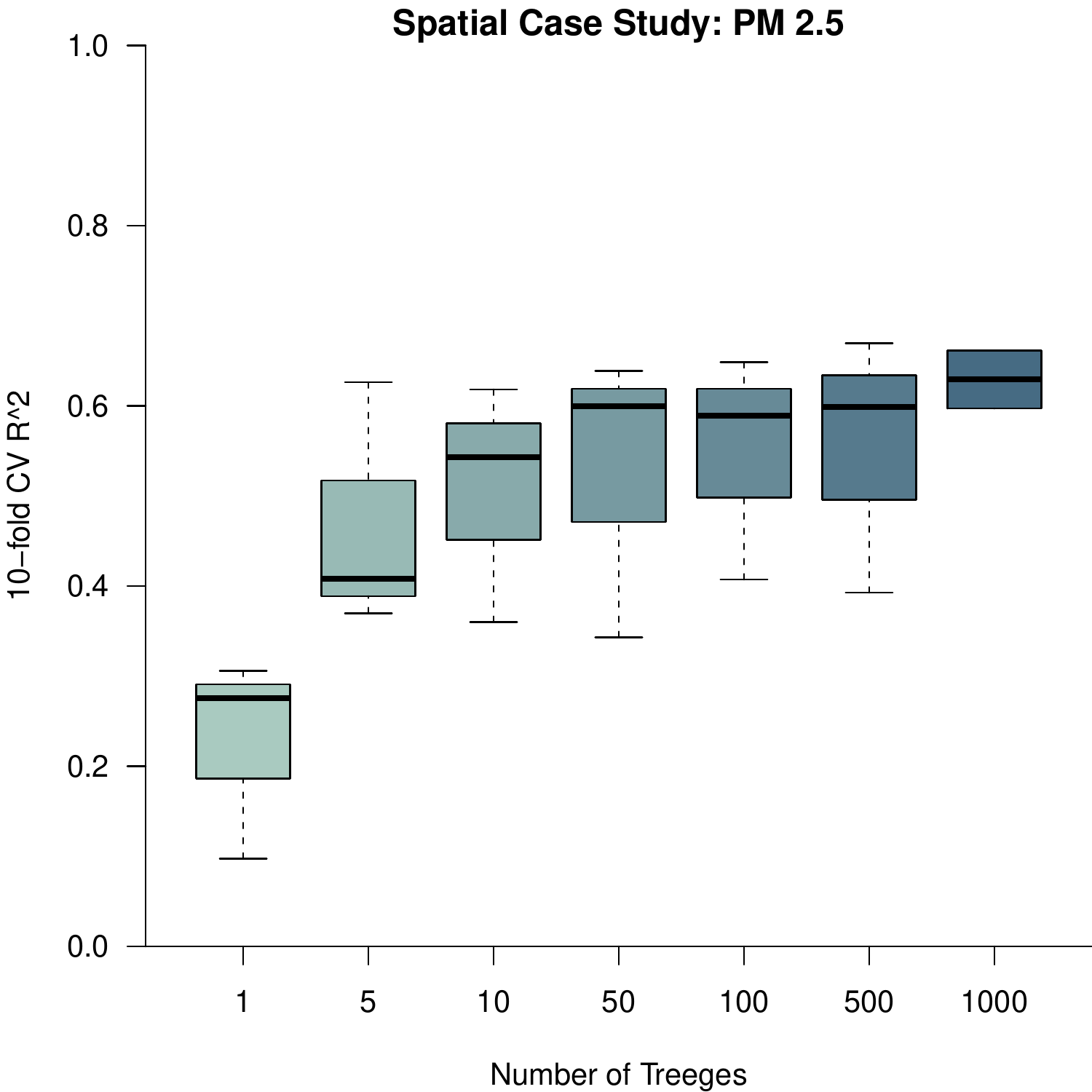}

\includegraphics[width=2.3in]{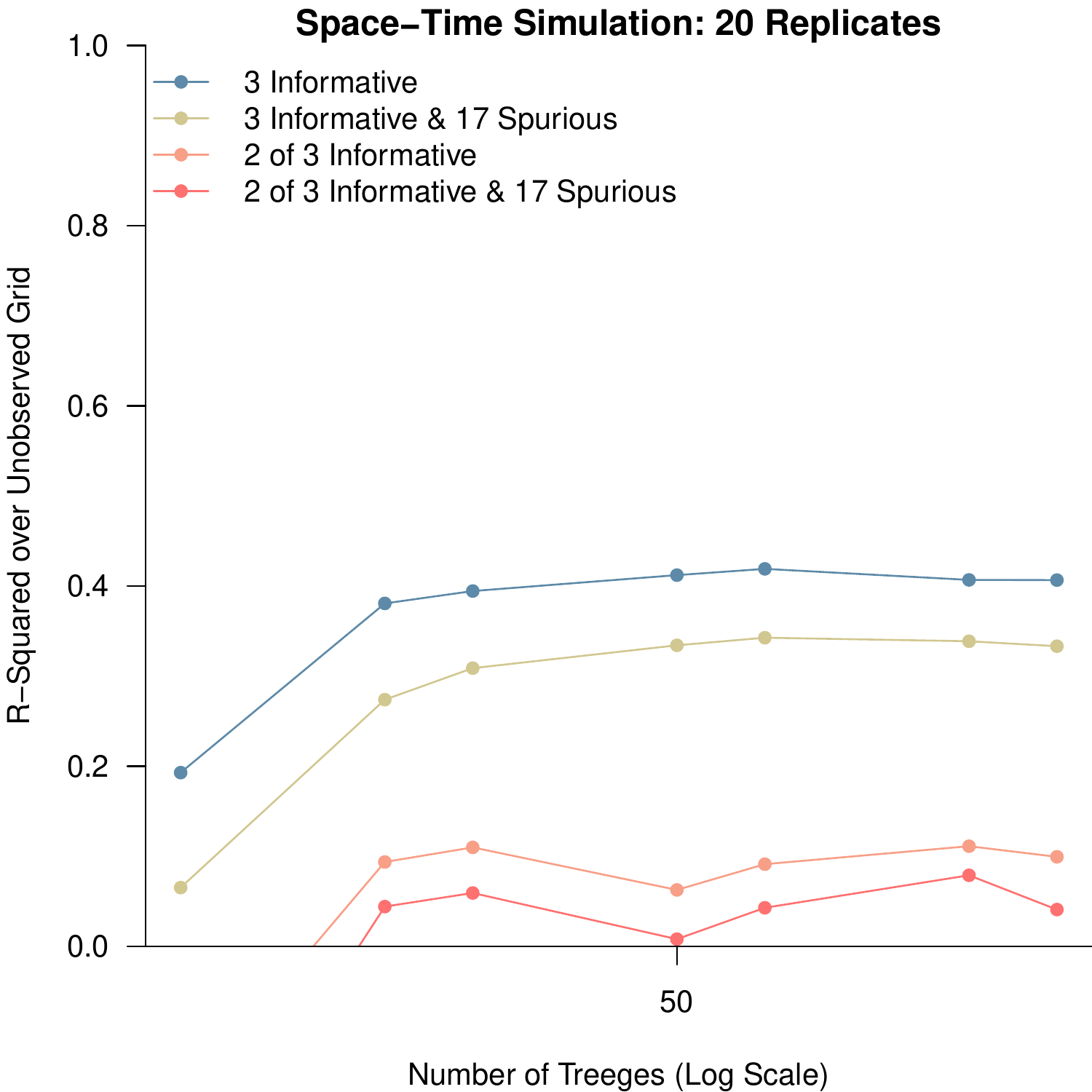}
\includegraphics[width=2.3in]{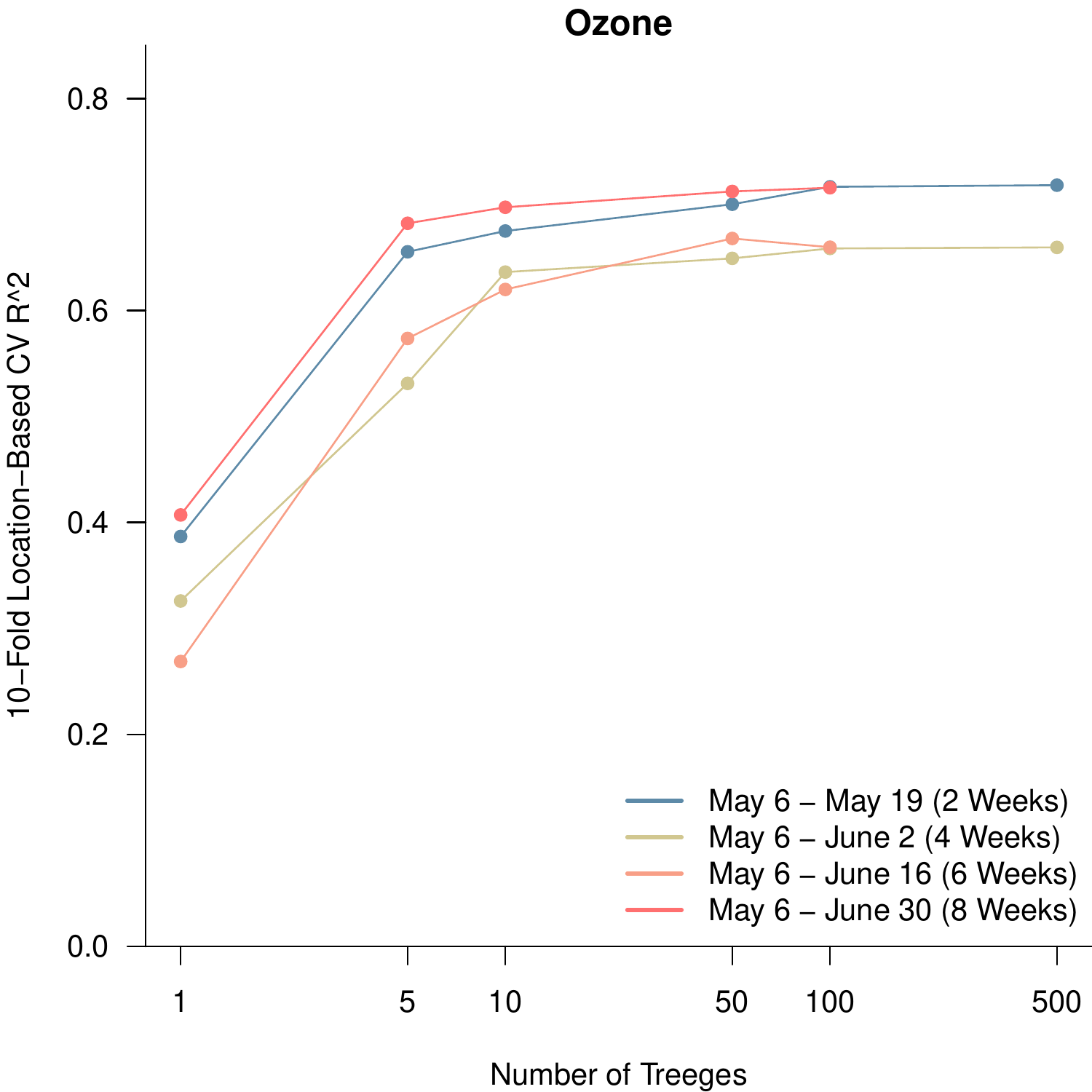}
\includegraphics[width=2.3in]{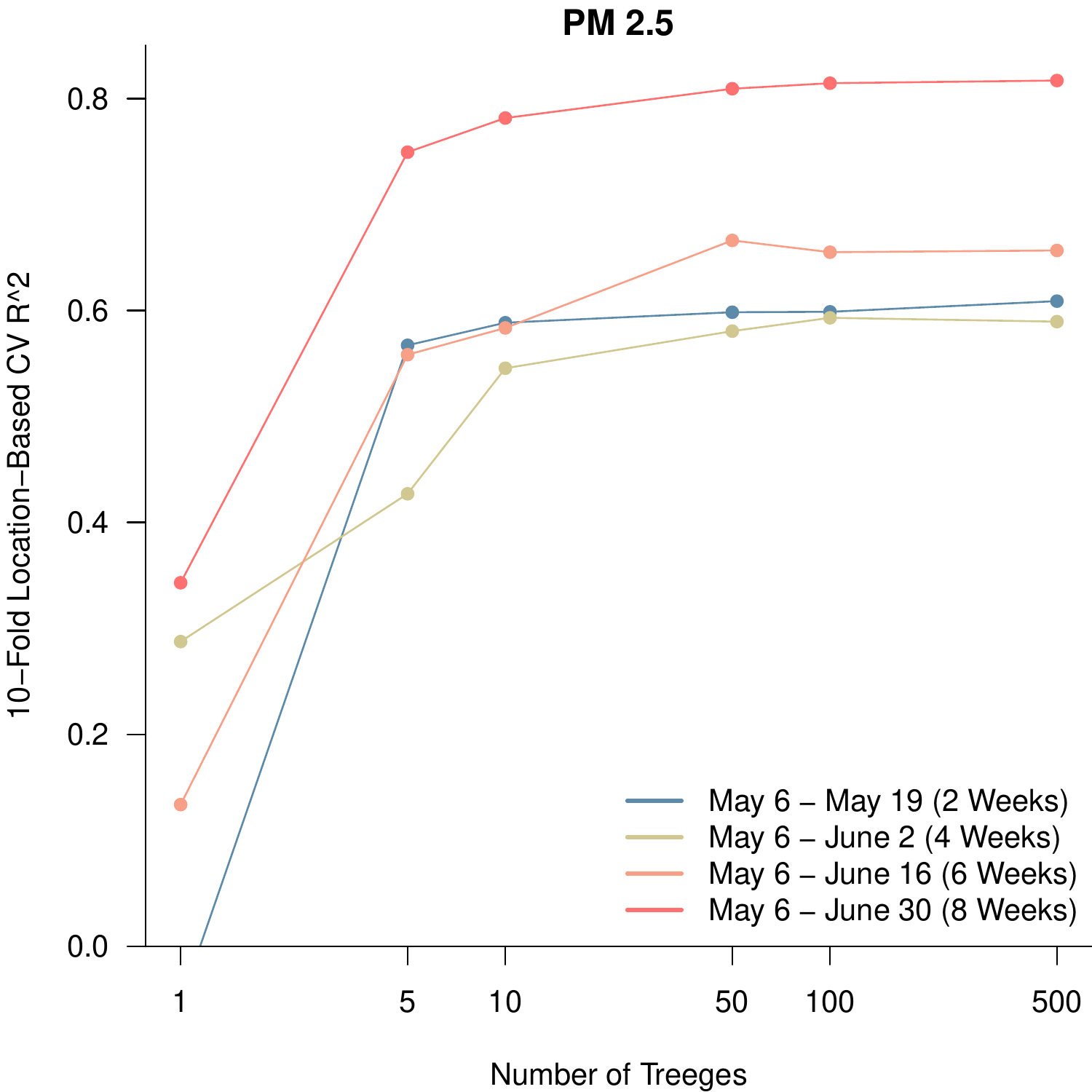}

  \caption{Tuning the Number of Treeges}
  \label{fig:tune_ntreege}
\end{figure}

\begin{figure}
\includegraphics[width=2.3in]{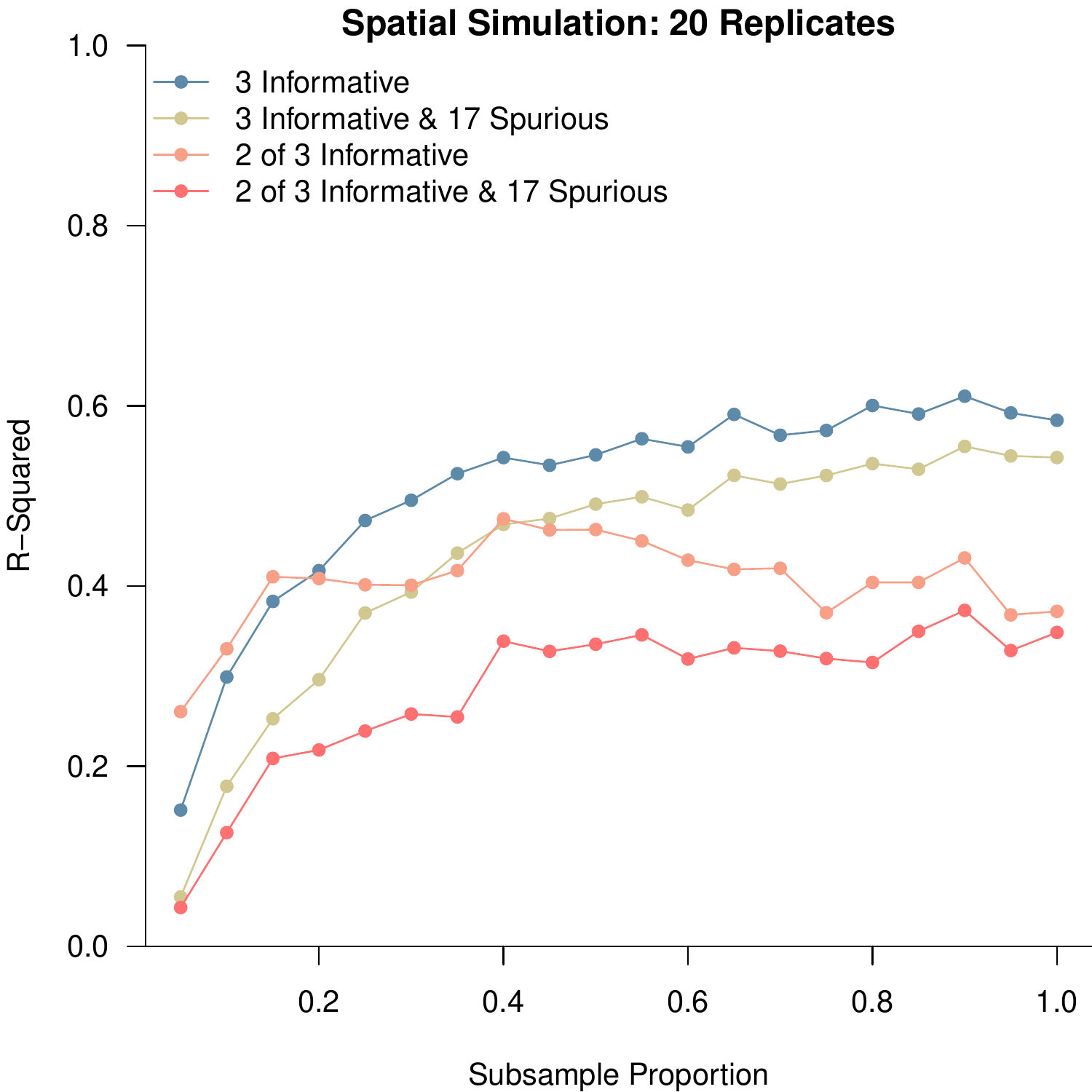}
\includegraphics[width=2.3in]{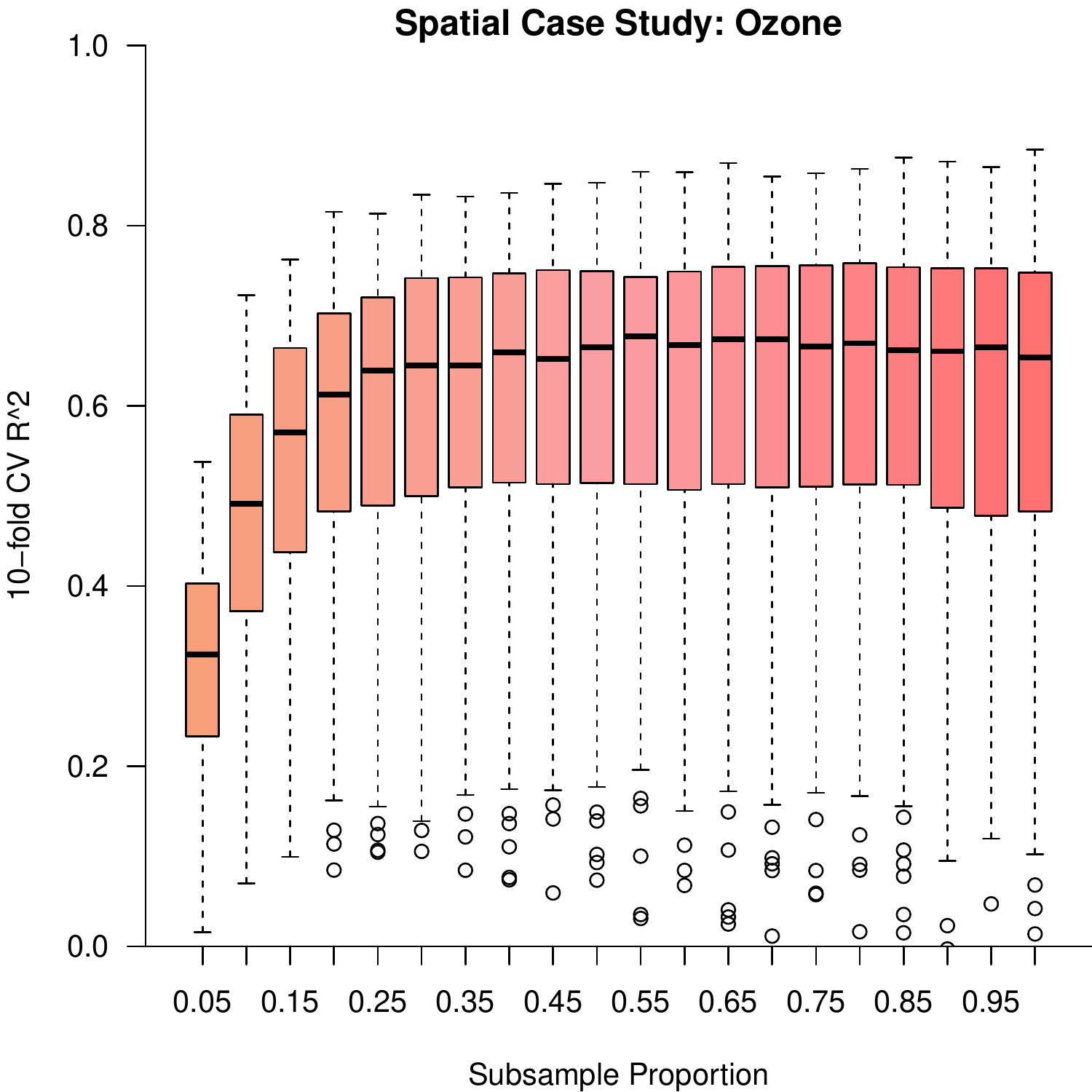}
\includegraphics[width=2.3in]{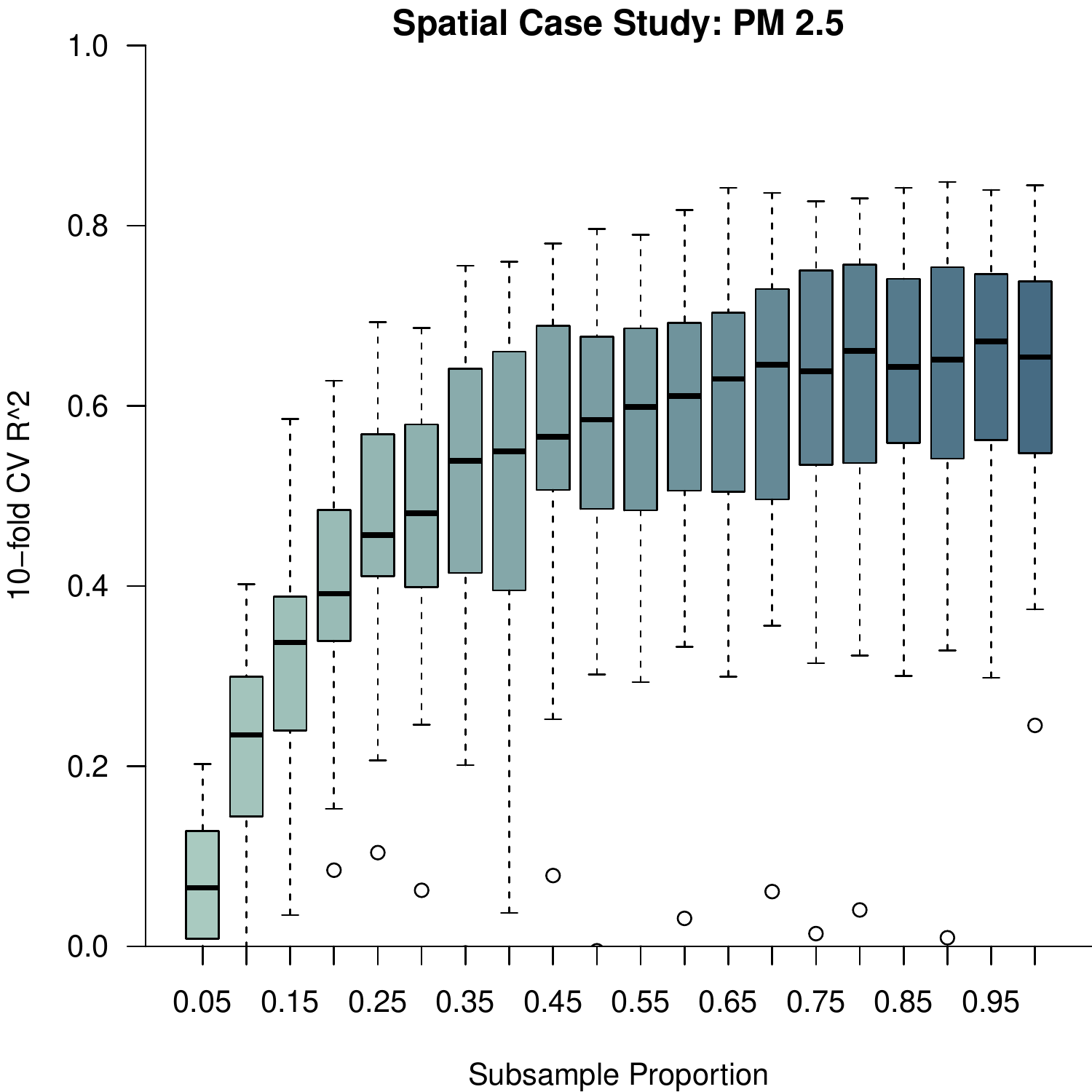}

\includegraphics[width=2.3in]{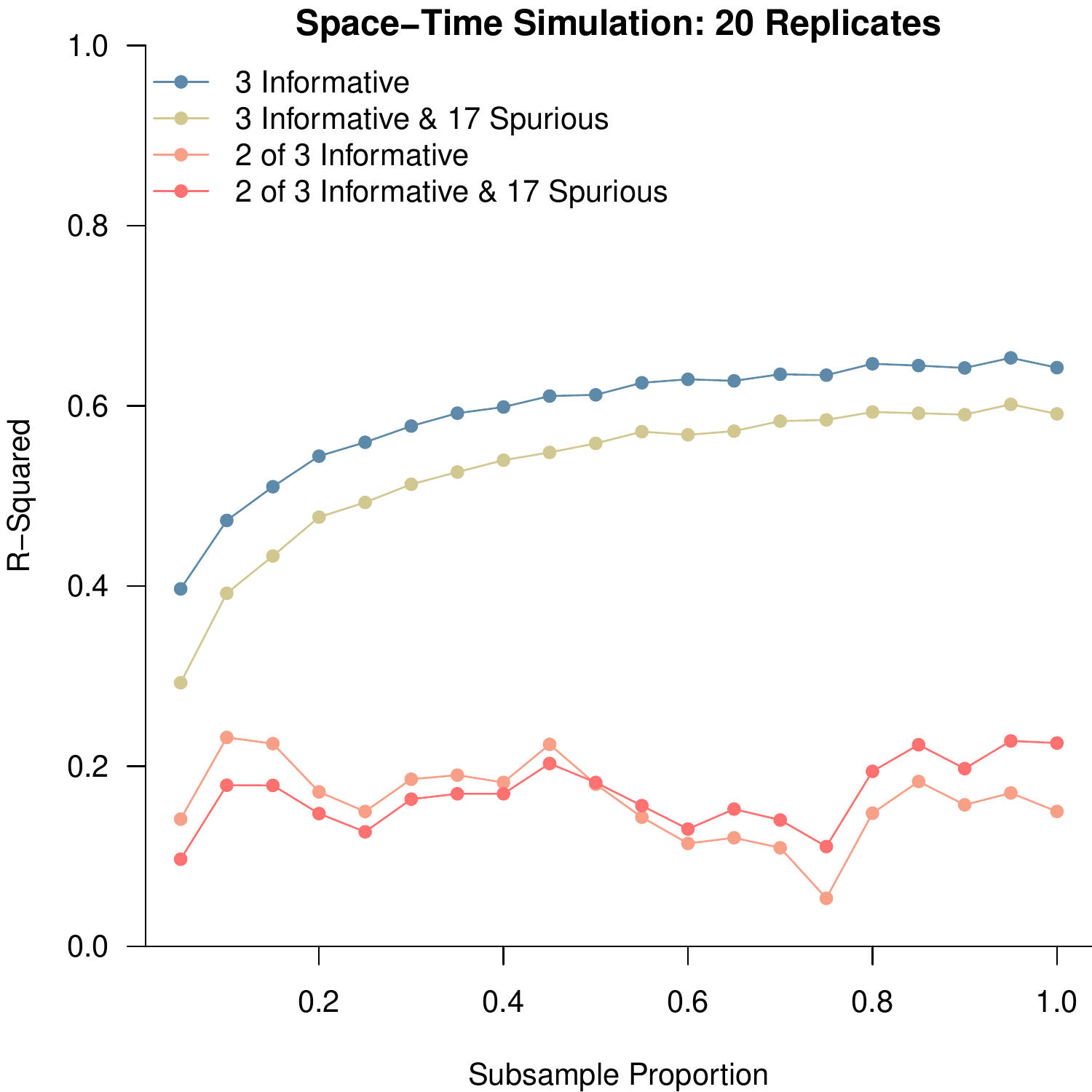}
\includegraphics[width=2.3in]{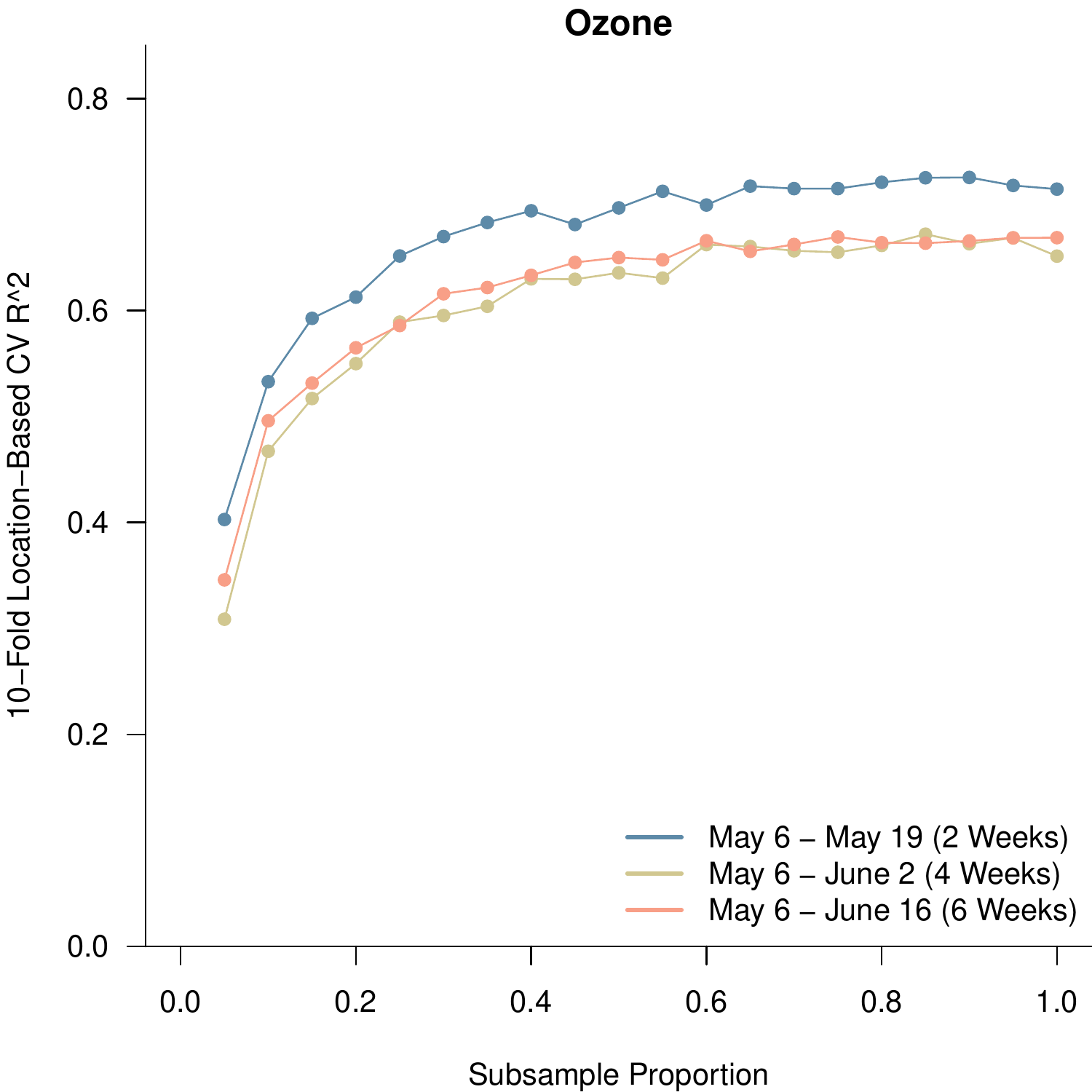}
\includegraphics[width=2.3in]{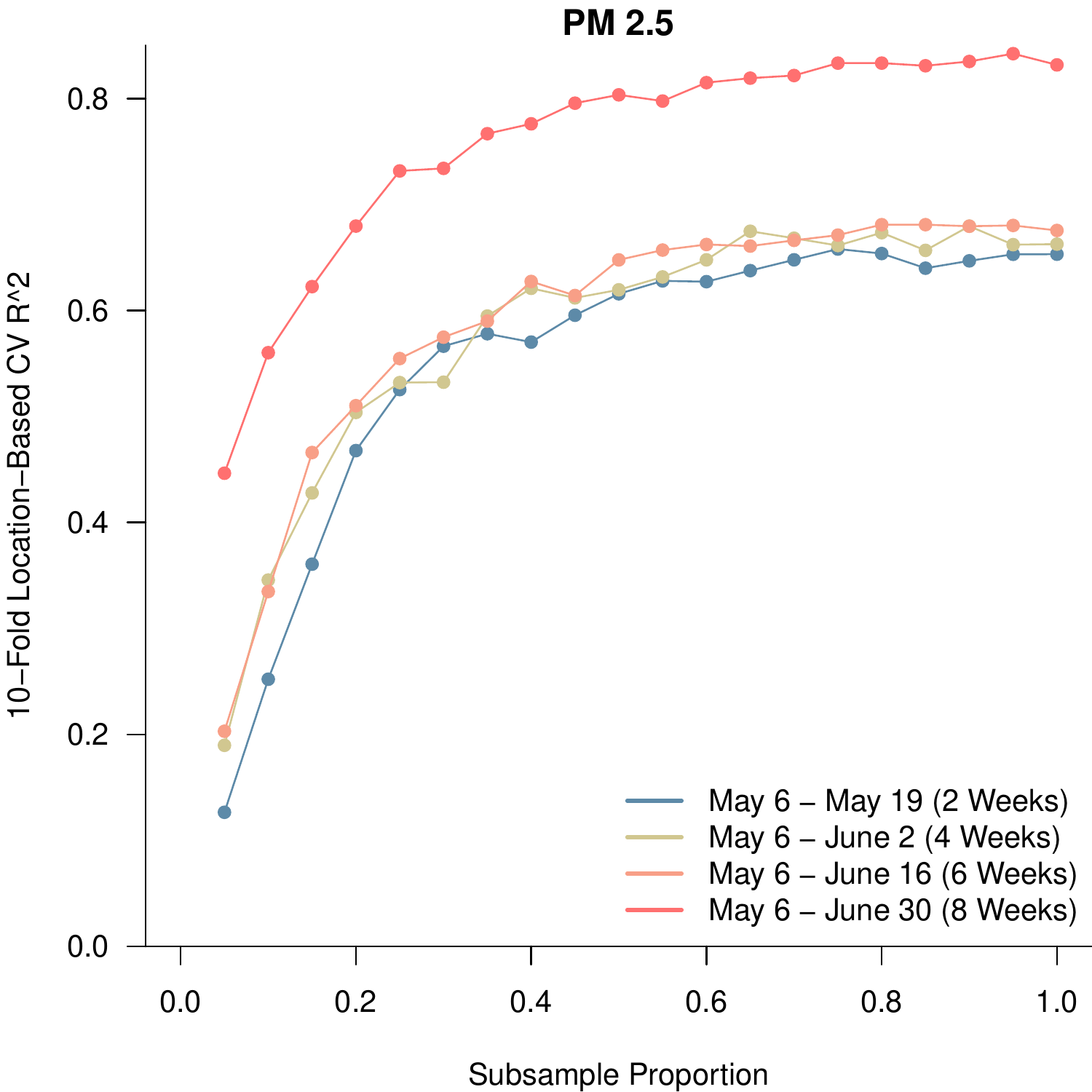}

  \caption{Tuning the Subsample Proportion}
  \label{fig:tune_subsamp}
\end{figure}

The two primary tunable parameters of treeging are the number of treeges and the subsample proportion. Figures~\ref{fig:tune_ntreege} and~\ref{fig:tune_subsamp} depict predictive $R^2$ (simulation) or $\hat{R}^2_{CV}$ (case studies) for a variety of values of these parameters. In all cases, there is a substantial improvement in predictive accuracy from 1 to 5 treeges, after which it generally increases until leveling off at 50 treeges. In some cases there may be evidence of subsequent improvement, but based on these results we set 50 as the default number of treeges. There is, of course, no guarantee this choice will be optimal for a particular problem. Computing time increases with the number of treeges, which may be an important consideration in the context of medium or large data. 

Similarly, predictive accuracy increases initially with subsample proportion $p$ and levels off. In both the spatial and space-time case studies, it seems to level off at a higher proportion for PM$_{2.5}$ than for ozone. In the spatial simulations there may be some evidence of an interesting divergence between scenarios in which all the informative covariates are observed (i and~ii) and those in which an informative covariate is withheld (iii and~iv). There appears to be a decrease in predictive accuracy for~iii and~iv for $p>0.75$, whereas accuracy continues to increase for~i and~ii. It may be that higher subsample proportions can exacerbate the effects of overfitting when the covariates are misspecified. These results reinforce our default choice of $p = 0.632$, motivated by~0.632 being the expected proportion of the observations that appear in a bootstrap sample. There is a computational discount on the order of $p^3$, which again may be relevant for medium or large data. 

\section{Large Data Set Approximations}
\label{sec:large_approx}

Treeging is computationally intensive. However, it is feasible whenever kriging is. The computational complexity of kriging is $o(n^3)$, making each treege approximately $o(p^3n^3)$ and a naive, sequential implementation of treeging $o(T p^3 n^3)$, where $p$ is the subsample proportion and $T$ is the number of treeges in the ensemble. Considerable gains can be made by parallelization, since treeges are fit independently. 

Additional gains can be made by exploiting sparsity in $\mathbf{X}_\tau(\mathbf{d})$ and spherical covariance matrices. The treege design matrix $\mathbf{X}_\tau(\mathbf{d})$ has one nonzero element per row, a 1 at the column corresponding to the leaf in which the covariates associated with that row reside. Consequently, only only $L$ of its $nL$ entries are nonzero, where $L$ is the number of leaves in the treege. The estimated covariance matrix $\hat{\Sigma}$ also tends to be sparse when a spherical covariance model is used, because the spherical covariance function decreases to 0 at a finite distance. Multiplying sparse matrices is generally substantially faster than dense matrices. The treeging R package uses the sparse matrix multiplication routines of the Basic Linear Algebra Subprograms (BLAS) via the Rcpp package~\citep{dongarra2002basic,eddelbuettel2011rcpp}. 

Treeging can also take advantage of strategies that have been developed to improve the scaleability of kriging or Bayesian Gaussian process regression, such as Nearest-Neighbor Gaussian Processes~\citep{datta2016hierarchical}. 

\section{Discussion}
\label{sec:disc}

Treeging is an effective, new tool for spatial and space-time prediction. By enriching the flexible mean structure of regression tree base learners with the covariance structures of Gaussian process prediction and ensembling them, it combines the strengths of the two major classes of space-time prediction algorithms. While it is not the optimal model in all scenarios, it is clearly the best performing overall, whereas kriging suffers when dependence is weak or the mean structure is misspecified and random forest suffers when the covariates are uninformative. 

Treeging offers clear advantages over alternatives that often rely on assumptions of smoothness and structural encoding of covariate effects~\citep{berrocal2010spatio, szpiro2010predicting}. These do not offer the black box benefits of ensembling strategies that automatically learn regression surfaces, requiring subject level knowledge on the number of basis functions and how to encode covariates. The generative adversarial networks that have  recently been applied to spatial interpolation problems~\citep{zhu2020spatial, manepalli2020, watson2020investigating, gao2020si} similarly require some level of subject knowledge, with expected performance of deep learning networks depending on a large number of tuning structures. 

Most other approaches for enriching tree-based ensembles for dependent data have limited their focus to the spatial case. Post-hoc kriging of random forest residuals attempts to correct the fit of a traditional random forest using spatial dependence, but does not incorporate dependence into the ensemble base learners~\citep{fayad2016regional,fox2020comparing}. Incorporating spatial distances between observations~\citep{hengl2018random} along with values of nearby observations~\citep{sekulic2020random} as features may incorporate spatial information into random forest but in a restrictive manner and may greatly increase the number of features in the random forest. Alternatively, fitting a spatially localized random forest~\citep{georganos2021geographical} allows for spatial heterogeneity but requires selecting weight and bandwidth parameters, which adds computational complexity and may not perform well in the presence of  nonstationarity. 

Treed Gaussian processes~(TGP)~\citep{gramacy2008bayesian} and RF-GLS~\citep{saha2021random} are more closely related. 
TGP is a Bayesian regression tree with a GP at each leaf. This provides a very flexible probability model, but practically does not scale even to small air pollution space-time data sets like the approximately 13,000 observation ozone case study considered here, and convergence of MCMC over the space of regression trees remains challenging. 
The ensemble-based model averaging of treeging may also be preferable to the Bayesian model averaging of TGP, since the latter assumes a single regression tree as the underlying mean structure. It is likely and the goal of future work to show that treeging approximates the MAP (maximum a posteriori) under a particular, restrictive version of TGP. 
RF-GLS is an extension of random forest into a GLS framework for spatial data, and \cite{saha2021random} provide some theoretical guarantees that may hold for treeging as well.

A number of extensions and improvements to treeging present themselves as useful avenues for future work. Naive ensembling could be extended to include a formalized notion of additivity and a formal optimization strategy, motivated by the well-known successes of boosting including on space-time air pollution problems~\citep{watson2019machine, reid2015spatiotemporal}. Localized subsampling may provide a useful alternative for treege construction, particularly in the presence of training data sampling bias. Random forest variable importance metrics may provide insight into the contribution of particular covariates within a treeging ensemble. One could also explore bootstrap strategies for uncertainty quantification and possibly investigate the role of out-of-bag (OOB) error as estimator of interpolation error. Some work in IID cases has been published making use of the OOB error for random forest uncertainty quantification~\citep{zhang2019random}. Finally, one could consider different flexible models for the mean structure.

\section{Supplementary Materials}

\begin{description}

\item[Appendix:] Supplemental details on simulations and models. (pdf)

\end{description}

\bigskip







\bibliographystyle{apalike}
\bibliography{refs}



\end{document}


\def\spacingset#1{\renewcommand{\baselinestretch}%
{#1}\small\normalsize} \spacingset{1}


\if0\blind
{
  \title{\bf Treeging: Appendix}
  \author{Gregory L. Watson$^1$\thanks{
    gwatson@ucla.edu}\hspace{.2cm}, Michael Jerrett$^2$ and
    Donatello Telesca$^1$ \\
    $^1$ Department of Biostatistics, University of California, Los~Angeles \\ 
    $^2$ Department of Environmental Health Sciences, University of California, Los~Angeles}
  \maketitle
} \fi

\if1\blind
{
  \bigskip
  \bigskip
  \bigskip
  \begin{center}
    {\LARGE\bf Title}
\end{center}
  \medskip
} \fi

\spacingset{1.45}
\section{Spatial Simulations}

The three informative and 17 spurious spatial simulation covariates were sampled from a Gaussian process with mean 0 and an independent covariance function. The spatial field, $Y(d)$, was sampled from a GP with a mean function $\mu_Y(d)$ depending on three covariates:
\begin{equation}
  \mu_Y(d) = \eta * \{X_1(d) + 1[X_2(d) \geq 0] - 1[X_2(d) < 0] + 3 [1 + \exp (-2X_3(d) + 3)]^{-1}\},
\end{equation}
where $a$ is the covariate effect size multiplier that scales up or down the strength of the covariate effects. An isotropic exponential covariance function was employed, $\Sigma(d_1, d_2) = \exp({- \lVert d_1 - d_2 \rVert / \nu })$, where $\nu$ is a parameter determines the strength of spatial dependence for points on the random field that are a distance of $\lVert d_1 - d_2 \rVert$ apart. 

The values of $\eta$ and $\nu$ were varied to investigate model performance under a variety of circumstances. The values of the effect size multipler, $\eta$, were $\{0, 0.1, 0.2 ..., 2 \}$, and $\nu$ varied from zero to 1.25, taking the following values,
\{0, 0.05, 0.1, 0.125, 0.15, 0.175, 0.2, 0.225, 0.25, 0.275, 0.3, 0.325, 0.35, 0.375, 0.4, 0.425, 0.45, 0.475, 0.5, 0.525, 0.55, 0.575, 0.6, 0.625, 0.65, 0.675, 0.7, 0.725, 0.75, 0.775, 0.8, 0.825, 0.85, 0.875, 0.9, 0.925, 0.95, 0.975, 1, 1.025, 1.05, 1.075, 1.1, 1.125, 1.15, 1.175, 1.2, 1.225, 1.25\}.

The unobserved grid of spatial locations at which $Y(d)$ was simulated was the cartesian product of 21 evenly spaced points begining at 0 and ending at 10. The observed data were simulated at 100 locations randomly selected with uniform probability $\{(s_1, s_2) : s_1, s_2 \in (0,10)\}$.

\section{Case Study Covariates}

  \small
  \label{tbl:Covariates}
  \begin{tabular}{ll}
    \hline
    Covariate  & Data Source \\
    \hline
Monitor Latitude	& U. S. Environmental Protection Agency \\
Monitor Longitude	& U. S. Environmental Protection Agency \\
Date$^1$	& U. S. Environmental Protection Agency \\
Elevation (m)	& National Digital Elevation Model \\
Boundary Layer Height (m)	& Rapid Update Cycle \\
Surface Pressure (Pa)	& Rapid Update Cycle \\
Relative Humidity (\%)	& Rapid Update Cycle \\
Temperature at 2 m ($^{\circ}$K)	& Rapid Update Cycle \\
U-Component of Wind Speed (m/s)	& Rapid Update Cycle \\
V-Component of Wind Speed (m/s)	& Rapid Update Cycle \\
Inverse Distance to Nearest Wildfire (m$^{-1}$)$^2$	& Fire Inventory from NCAR v1.5 \\
Annual Average Traffic within 1 km	& Dynamap 2000, TeleAtlas \\
Agricultural Land Use within 1 km (\%)	& 2006 National Land Cover Database \\
Urban Land Use within 1 km (\%)	& 2006 National Land Cover Database \\
Vegetation Land Use within 1 km (\%)	& 2006 National Land Cover Database \\
Normalized Difference Vegetation Index	& Landsat Data \\
Nitrogen Dioxide (log molecules/cm$^2$)	& Ozone Monitoring Instrument Satellite \\
WRF-Chem PM$_{2.5}$  (log kg/day))$^3$ & WRF-Chem \\
WRF-Chem Ozone (log 8 Hour Maximum)$^4$)	& WRF-Chem  \\
    \hline
  \end{tabular}
\\

\noindent $^1$ Date was included for space-time simulations only.

\noindent $^2$ Proximity to wildfires was incorporated as inverse distance to the nearest wildfire. See~\cite{watson2019machine} for details. 

\noindent $^3$ Only included as covariate for PM$_{2.5}$ case studies. 

\noindent $^4$ Only included as covariate for ozone case studies.

\section{Spherical Covariance Estimation}

The default covariance estimation procedure for treeging fits a spherical variogram to the empirical variogram of the spatial or temporal residuals. For space-time treeging, we assume separability and separately fit spatial and temporal spherical covariances, multiplying them together to estimate the space-time covariance. The residuals are the difference between the fitted and observed training outcomes, i.e., 
\begin{equation}
    \mathbf{e}(\mathbf{d}) = \mathbf{y}(\mathbf{d}) - \hat{\mathbf{y}}(\mathbf{d}),
\end{equation}
where $\hat{\mathbf{y}}(\mathbf{d}) = T(\mathbf{X}(\mathbf{d}), \boldsymbol \theta)$ for treeging and $\hat{\mathbf{y}}(\mathbf{d}) = \mathbf{X}(\mathbf{d}) \boldsymbol \beta$ for  kriging. The empirical variogram is 
\begin{equation}
    \hat{\gamma}(h \pm \delta) = \frac{1}{2 \lvert N(h \pm \delta) \rvert} \sum_{\lvert d_i - d_j \rvert \leq \delta} \lvert y(d_i) - y(d_j) \rvert ^2,
\end{equation}
and the spherical variogram is
\begin{equation}
\gamma(h,r,s,a) = 
\begin{cases} 
0 & h = 0\\
a + (s - a) \left(\frac{3h}{2r} - \frac{h^3}{2r^3}\right) & 0 < h \leq r\\
s & h > r, 
\end{cases}
\end{equation}
and spherical covariance function is
\begin{equation}
C(h,r,s,a) = 
\begin{cases} 
s & h = 0\\
(s - a) \left(1 - \frac{3h}{2r} + \frac{h^3}{2r^3}\right) & 0 < h \leq r\\
0 & h > r.
\end{cases}
\end{equation}

\section{Space-Time Simulations}

Covariates for the space-time simulations were generated from a Gaussian process (GP) with mean 0 and a separable covariance function $\Sigma_X(d) = \Sigma_X(s)\Sigma_X(t)$, where $d = (s,t)$. Exponential covariance functions with randomly selected range parameters were used for $\Sigma_X(s)$ and $\Sigma_X(t)$.  The random field mean 
$\mu_Y[\mathbf{X}(d)]$ was a nonlinear function of the first 3 covariates, 
\begin{equation}
   \mu_Y[\mathbf{X}(d)] = \eta \{\tfrac{1}{2}X_1(d)
     + \tfrac{3}{4} [X_2(d)  \leq 0.1] - \tfrac{3}{4}  [X_2(d) > 0.1]
     + 2  \{1 + \exp[-2 X_3(d)]\}^{-1}
     + X_i(d)Xj(d) \},
\end{equation}
where $\eta$ is the covariate effect size multiplier that can be scaled up or down to simulate stronger or weaker covariate effects, and $i$ and $j$ are 2 of the 3 informative covariates selected at random. Values of the effect size multiplier $\eta$ were $\{0, 0.2, 0.4, ... , 2\}$. The additional covariates were included as spurious covariates in scenarios $B$ and $D$ to assess the robustness of prediction models.

A separable covariance function was used for $Y(d)$ with independent exponential spatial and temporal covariance functions. The ranges were varied across the following ranges to investigate the effects of varying levels of spatial and temporal dependence. Values for the spatial covariance range were $2 - \sqrt{u}$ for $u \in \{0, 0.25, 0.5, ..., 4\}$. Values for the temporal covariance range were $\{0, 25., 5, 7.5, 10\}$. 

The unobserved grid of spatial locations at which $Y(d)$ was simulated was the cartesian product of 11 evenly spaced points begining at 0 and ending at 10. The observed data were simulated at 40 spatial locations randomly selected with uniform probability $\{(s_1, s_2) : s_1, s_2 \in (0,10)\}$. Both the observed and unobserved data were simulated over 30 consecutive timepoints, yielding a total of 1,200 observed data points and 3,630.

\bibliographystyle{apalike}
\bibliography{refs}